\documentclass{ws-ijbc}
\usepackage{ws-rotating}     
\usepackage{graphicx}
\usepackage{epstopdf}
\usepackage{natbib}
\usepackage{xcolor}
\usepackage{soul}
\usepackage[symbols,nogroupskip,nonumberlist,nopostdot=false,abbreviations]{glossaries-extra}
\newabbreviation{kde}{KDE}{Kernel Density Estimation}
\newabbreviation{hde}{HDE}{Histogram Density Estimation}
 \newabbreviation{iid}{i.i.d.}{independently and identically distributed}
\glsxtrnewsymbol[description={Mean Square Error}]{mse}{\ensuremath{\operatorname{\text{MSE}}}}
\glsxtrnewsymbol[description={Upper Bound}]{ub}{\ensuremath{\operatorname{\text{UB}}}}
\glsxtrnewsymbol[description={Bias of the estimator}]{bias}{\ensuremath{\operatorname{\text{bias}}}}
\glsxtrnewsymbol[description={Variance}]{var}{\ensuremath{\operatorname{\text{Var}}}}
\glsxtrnewsymbol[description={Gauss error function}]{erf}{\ensuremath{\operatorname{\text{erf}}}}
\glsxtrnewsymbol[description={Forward orbit of a dynamical system $f$}]{orbit}{\ensuremath{\operatorname{O^+}}}
\glsxtrnewsymbol[description={Frobenius-Perron operator}]{fp}{\ensuremath{\cal P}}
\glsxtrnewsymbol[description={Map $f:\Omega \rightarrow \Omega$ such that 
 $x \mapsto f(x)$}]{f}{\ensuremath{f}}

 \glsxtrnewsymbol[description={State space of the map $f$}]{Omg}{\ensuremath{\Omega}}
 \glsxtrnewsymbol[description={Probability Density Function of random variable $X$}]{pdf}{\ensuremath{\rho_X}}
 \glsxtrnewsymbol[description={Estimated Probability Density Function with parameter $\theta$}]{epdf}{\ensuremath{\overline{\rho_{\theta}}}}
 \glsxtrnewsymbol[description={Matrix approximation to the Frobenius-Perron operator}]{eFP}{\ensuremath{\mathbf{P}}}

 \glsxtrnewsymbol[description={$i^{th}$ bin or box such that $\mathcal{B}=\{B_i\}_{i=1}^K$, $K>0$ is a finite topological cover of $\Omega$}]{Bi}{\ensuremath{B_i}}

  \glsxtrnewsymbol[description={Kernel in a \gls{kde}}]{kernel}{\ensuremath{\operatorname{\mathcal{K}}}}

   \glsxtrnewsymbol[description={Random Variable with the state space $\Omega$}]{rv}{\ensuremath{X}}

\glsxtrnewsymbol[description={Transformed random variable such that $X'=f(X)$}]{trv}{\ensuremath{X'}}

\glsxtrnewsymbol[description={Standard deviation of a distribution}]{sd}{\ensuremath{\sigma}}

\glsxtrnewsymbol[description={Number of bins in a histogram}]{k}{\ensuremath{K}}

\glsxtrnewsymbol[description={Bandwidth of a Kernel}]{del}{\ensuremath{\delta}}
\usepackage[capitalize,noabbrev]{cleveref}
\makeglossaries

\begin{document}


\markboth{S.~Surasinghe, J.~Fish \and E.~M.~Bollt}{Learning Transfer Operators by Kernel Density Estimation}

\title{LEARNING TRANSFER OPERATORS BY KERNEL DENSITY ESTIMATION}

\author{{Sudam Surasinghe} $^{1,2,3}$*, {Jeremie Fish} $^{1,\dagger}$ and {Erik M.~Bollt} $^{1,\ddagger}$ }

\address{ $^{1}$ Department of Ecology and Evolutionary Biology, Yale University, New Haven, CT, 06520 USA\\
$^{2}$ Public Health Modeling Unit, Yale School of Public Health, New Haven, CT 06510 USA\\ 
$^{3}$ Clarkson Center for Complex Systems Science, Department of Electrical and Computer Engineering, Clarkson University, 8 Clarkson Ave, Potsdam, New York 13699, USA\\
* \email{sudam.surasinghe@yale.edu} $^{\dagger}$ \email{fishja@clarkson.edu} $^{\ddagger}$  \email{ebollt@clarkson.edu}}


\maketitle


\begin{abstract}
Inference of transfer operators from data is often formulated as a classical problem that hinges on the Ulam method. The conventional description, known as the Ulam-Galerkin method, involves projecting onto basis functions represented as characteristic functions supported over a fine grid of rectangles. From this perspective, the Ulam-Galerkin approach can be interpreted as density estimation using the histogram method. In this study, we recast the problem within the framework of statistical density estimation. This alternative perspective allows for an explicit and rigorous analysis of \gls{bias} and variance, thereby facilitating a discussion on the mean square error. Through comprehensive examples utilizing the logistic map and a Markov map, we demonstrate the validity and effectiveness of this approach in estimating the eigenvectors of the Frobenius-Perron operator. We compare the performance of \gls{hde} and \gls{kde} methods and find that \gls{kde} generally outperforms \gls{hde} in terms of accuracy. However, it is important to note that \gls{kde} exhibits limitations around boundary points and jumps. Based on our research findings, we suggest the possibility of incorporating other density estimation methods into this field and propose future investigations into the application of \gls{kde}-based estimation for high-dimensional maps. These findings provide valuable insights for researchers and practitioners working on estimating the Frobenius-Perron operator and highlight the potential of density estimation techniques in this area of study.
\end{abstract}


\section{Introduction}
Transfer operators play a vital role in the global analysis of dynamical systems. The availability of large amounts of data from dynamical systems drives the popularity of these operators in data-driven analysis methods for complex systems. Therefore, accurately estimating the transfer operators from data is crucial for successful global analysis. The Frobenius-Perron operator is one such popular operator used for global analysis of dynamical systems, and the Ulam method \cite{ulam1960collection} is the most popular method to estimate it\cite{hunt1998unique,ermann2010ulam,junge2009discretization,hunt2000attractors,froyland2013estimating}. However, as we point out here, there are tremendous opportunities to recast this problem as one of density estimation as it is called in the statistics literature. This approach offers a well-developed analysis of variance and \gls{bias}, enabling discussions of mean squared error. This language has been overlooked in the dynamical systems community. Therefore, we introduce the probability density estimation viewpoint  to estimating the Frobenius-Perron operator, and with it, the rich analysis already developed in other mathematical communities and methods, notably kernel estimation which as it turns out is provably more efficient than the histogram methods used in the standard Ulam method. Also, note that while kernel regression methods are commonly used in operator estimation problems, such as Koopman operators (\cite{klus2018kernel,klus2019kernel,klus2018data, mollenhauer2020singular}), kernel density estimation is a distinct technique.

A Frobenius-Perron operator evolves the density of ensembles of initial conditions of a dynamical system forward in time. This statement can also be re-interpreted in a Bayesian framework. Hence, we argue that we have essentially a problem of density estimation for the conditional probability density that is generally described as the Frobenius-Perron operator. Therefore, the classical Ulam method is essentially a histogram method for the estimation of this conditional density function by simple nonparametric means. Many, including one of the authors of this work, have described the approach as a projection onto basis functions as characteristic functions. In these terms, it was described as Ulam-Galerkin's method \cite{bollt2013applied}, which covers many of the analyses of convergence since the original conjecture \cite{ CHIU1992291, Boyarsky1997LawsOC, froyland2001, GUDER1997525}.

However, it is generally understood that histograms, while easy to describe, are a primitive variant among the approaches available to the problem of nonparametric density estimation.  It has been said by Tukey \cite{tukey1961curves,tukey1981graphical}, that the appearance of the traditional histogram is blocky, and difficult to balance smoothing, bandwidth, \gls{bias}, and variance.  Even in two dimensions, ``blocky" variability of sampling, and details such as choosing an appropriate orientation of the grid, become problematic.  There are, however, more suitable methods are reviewed here, especially \glsdesc{kde} (\gls{kde}) which has many nice smoothing, analytic, and convergence properties. Additionally, density estimation is shifted from a question of density in space to an expectation of points.  Note also that the k-nearest neighbor (kNN) methods also have many of these advantages, but kernel methods allow for better tuning of smoothing parameters and good convergence statistics.

It is argued in \cite{dehnad_1987}, that the argument for \gls{kde} instead of a simple histogram method becomes stronger in more than one dimension, due to difficulties not only in histogram box size (bandwidth) but now also, in orientation and origin location that generally lead to a blocky appearance that becomes more difficult to interpret the joint and conditional probabilities.  Tukey asserted \cite{tukey1981graphical}, "...it is difficult to do well with bins in as few as two dimensions.  Clearly, bins are for the birds!"

In this article, we aim to provide a comprehensive analysis of the density estimation approach for estimating the Frobenius-Perron operator. We begin by discussing the standard theory for the operator and the classical Ulam-Galerkin method in \cref{sec:StdThr}. Using Bayesian theory (in \cref{Sec:BayInter}), we then analyze the Ulam method as a histogram-based density estimation technique and argue for the use of a standard density estimation approach for estimating the Frobenius-Perron operator. Then, in \cref{Sec:TheoryDen}, we present a theoretical analysis that highlights the advantages of \gls{kde} over \gls{hde} (note that the Ulam method is based on \gls{hde}), particularly in terms of variance and bias. We also conduct a comprehensive examination of the convergence properties of kernel density estimation. Furthermore, we apply our theoretical analysis to the well-known chaotic logistic map example in \cref{sec:rest} and use the Markov map example to validate the reinterpretation of the Ulam method as a density estimation approach. By comparing the accuracy of the \gls{kde} approach with the Ulam method, which we have shown to be a form of \gls{hde}, we demonstrate the higher accuracy and performance of the former in estimating the Frobenius-Perron operator.

Overall, our study highlights the benefits of adopting a standard density estimation approach for estimating transfer operators, particularly in the context of data-driven analysis methods for complex systems. By recasting the problem of estimating the Frobenius-Perron operator as a density estimation problem, we can leverage the well-developed theory and methods in the statistics literature to improve the accuracy and efficiency of the estimation process. In order to maintain consistency with the standard notation used in the statistics literature \cite{silverman_1999,scott_2015}, we will adopt the symbols commonly used in density estimation theory. \cref{tab:symbols} shows the symbols that will be used throughout this article. 
\begin{table}[h!]
\centering 
\tbl{List of Symbols: Throughout the article, we will use these symbols to denote the relevant quantities and functions.}{
\begin{tabular}{p{4cm}p{12cm}}
\toprule
\textbf{Symbol} & \textbf{Description}\\
\colrule
$\gls{Omg}$ & \glsdesc{Omg} \\
$\gls{f}$ & \glsdesc{f} \\
$\gls{orbit}$ & \glsdesc{orbit} \\
$\gls{rv}$ & \glsdesc{rv} \\
$\gls{trv}$ & \glsdesc{trv} \\
$\gls{sd}$ & \glsdesc{sd} \\
$\gls{k}$ & \glsdesc{k} \\
$\gls{del}$ & \glsdesc{del} \\
\gls{pdf} & \glsdesc{pdf} \\
\gls{epdf} & \glsdesc{epdf} \\
\gls{fp} & \glsdesc{fp} \\
\gls{eFP} & \glsdesc{eFP} \\
\gls{Bi} & \glsdesc{Bi} \\
\gls{kernel} & \glsdesc{kernel} \\
\gls{mse} & \glsdesc{mse} \\
\gls{ub} & \glsdesc{ub} \\
\gls{bias} & \glsdesc{bias} \\
\gls{var} & \glsdesc{var} \\
\gls{erf} & \glsdesc{erf} \\
\botrule
\end{tabular}\label{tab:symbols}
}
\end{table}
\section{Frobenius-Perron and the Classical Ulam-Galerkin Method for Estimation}\label{sec:StdThr}
First, we briefly review a standard discussion of Frobenius-Perron operators for deterministic and then random maps, and flows are covered.
Assuming a map, 
\begin{eqnarray}\label{map}
f:\Omega &\rightarrow& \Omega, \nonumber \\
 x &\mapsto& f(x),
\end{eqnarray}
the forward orbit of an initial condition $x$, $\gls{orbit}(x)=\{x,f(x),f^2(x),...\}$ is a fundamental object in the study of dynamical systems. However, if we consider an ensemble of many initial conditions, that are distributed by $\rho\in L^1(\Omega)$, and we assume $\gls{f}$ is a nonsingular transformation and measurable relative to $(\Omega,\mathbb{ B},\mu)$ on a Borel sigma-algebra of measurable sets $ \mathbb{ B} \subset \Omega$ with measure $\mu$, then the Frobenius-Perron operator that describes the orbit of ensembles (here we are following the notation from \cite{bollt2013applied}, and comparable to \cite{lasota1982exact}).  The linear map, ${\cal P}_f:L^1(\Omega)\rightarrow L^1(\Omega)$, follows the discrete continuity equation, 
\begin{equation}
    \int_B \rho_{n+1} d\mu=\int_{f^{-1}(B)} \rho_n d\mu, \mbox{ for any } B\in \mathbb{ B}, \label{mapnum1}
\end{equation}
where $\rho_n$ denotes the $n^{\text{th}}$ iterate of the initial density.
For differentiable maps, this simplifies, 
\begin{equation}
{\cal P}_f[\rho](x)=\sum_{y:x=f(y)} \frac{\rho(y)}{|Df(y)|},
\end{equation}
where if $f(y)$ is a univariate function, then $|Df(y)|=|f'(y)|$ is the absolute value of the derivative, or it is the determinant of the Jacobian (matrix) if multivariate.
The following equivalent form is relevant for our purposes here,
\begin{equation}\label{delta}
    {\cal P}_f[\rho](x)=\int_{\Omega} \delta(x-f(y))\rho(y) dy,
\end{equation}
in terms of the delta function.  Also, we have specialized to Lebesgue measure on $\Omega$ from this point forward.

The Ulam-Galerkin method is a way to estimate the action of the Frobenius-Perron operator, given a (fine) finite topological cover of $\Omega$ by (usually rectangles, or boxes, or triangles, or other simple spatial elements) ${\cal B}=\{B_i\}_{i=1}^K$, $K>0$.  The matrix approximation $\gls{eFP}$ is defined as follows:
\begin{equation}\label{est1}
    \gls{eFP}_{i,j}=\frac{m(B_i\cap f^{-1}(B_j))}{m(B_i)},
\end{equation}
where $m(B)=\int_B dx$ represents the Lebesgue measure. In fact, a simple estimate of this $K\times K$ matrix $\gls{eFP}$ follows if a large collection of input, output pairs are available, $(x_n,x_{n+1})$, as examples of $x\in B_i\cap  f^{-1}(B_j)$ perhaps derived from a long orbit that samples the space.  Note that $f^{-1}$ denotes the pre-image of $f$ which may not be one-one. This estimate is given by:
\begin{equation}\label{est2}
    \gls{eFP}_{i,j}\sim \frac{\# x_n\in B_i, x_{n+1}\in B_j}{\# x_n\in B_i}.
\end{equation}
Notice that the ``$\cap$" notation for the intersection of sets is coincident with the ``$,$" notation for ``and" which denotes both events occur.  This is useful for reinterpretation by a Bayesian discussion in the next section.  Under the above construction, it can easily be seen that $\gls{eFP}$ is a stochastic matrix, which therefore has a leading eigenvalue of $1$ and, if simple,  a dominant eigenvector which describes the steady state of the corresponding Markov chain.

The original Ulam-conjecture \cite{ulam1960collection} described that, in a limit of a refining partition $\{B_i\}$, the dominant eigenvector of the discrete state Markov chain  converges to the invariant density of the original dynamical system.  This conjecture was first proved by Li \cite{li1976finite} under the hypothesis of bounded variation of one-dimensional maps, providing weak convergence.

An estimate such as \cref{est1,est2} has previously been called a Ulam-Galerkin estimate ~\cite{bollt2013applied,ma2013relatively},  a description which was made to intentionally separate the concept of the limit of long time iteration, from short time considerations.  The phrase Galerkin is stated in terms of projection of the action of the operator onto a basis of characteristic functions $\{\xi_{B_i}(x)\}$, supported over the grid elements, $\{ B_i \}$,
\begin{equation}\label{eq:indicFac}
\xi_{B_i}(x)=
\begin{cases}1,\mbox{if} \ x \in B_i \\ 0, \ \mbox{otherwise} \end{cases}.
\end{equation}
Then the Ulam-Galerkin estimate formally describes a projection, $R:L^2(\Omega) \rightarrow \Delta_K$, for a finite linear subspace $ \Delta_K \subset L^2(\Omega)$, that is spanned by the collection of characteristic functions over the grid elements. Now, consider an operator $\mathcal{\bar{P}}_K(f): \Delta_K \rightarrow \Delta_K$ define by 
\begin{equation}
    \mathcal{\bar{P}}_K(f)[\xi_{B_i}](x)= \sum_{j=1}^K \mathbf{P}_{ij} \xi_{B_j}(x).
\end{equation}
Ulam proposed \cite{ulam1960collection} that if the transfer operator $\mathcal{P}_f$ has a unique fixed point (up to linear independence), then the sequence of fixed points $\rho_K$ of $\mathcal{\bar{P}}_K(f)$ should converge to a fixed point of the operator $\mathcal{P}_f$ as $K \rightarrow \infty$ \cite{li1976finite}. Notice this description is in terms of $L^2(\Omega)$, in order that an inner product structure makes sense, and then 
\begin{equation}\label{est3}
\gls{eFP}_{i,j}=\frac{m\big(\xi_{B_i},\xi_{f^{-1}(B_j)}\big)}{\|\xi_{B_i}\|}=\frac{\int_{\Omega}  \xi_{B_i}(x)\xi_{f^{-1}(B_j)}(x) dx }{\int_{\Omega} \xi_{B_i}(x)^2 dx} .
\end{equation}
If considering this finite rank transition in finite time we worry only about the estimation of transitions using the basis functions as discussed in \cite{bollt2013applied, bollt2002manifold}.  Infinite-time questions are clearly more nuanced which is why the Ulam conjecture remained unproven for almost twenty years.  Our Bayesian discussion will likewise avoid the same.  

In the more general case of a random dynamical system, \cref{mapnum1} is recast,
\begin{eqnarray}\label{map2}
x_{n+1}=f(x_n)+s_n,
\end{eqnarray}
which describes a deterministic part $f$ together with a stochastic ``kick" $s$ which we assume is identically independently distributed by $s\sim \nu$.  Consequently, the kernel integral form of the Frobenius-Perron transfer operator becomes,
\begin{equation}\label{delta2}
    {\cal P}_f[\rho](x)=\int_{\Omega} \nu(x-f(y))\rho(y) dy,
\end{equation}
which we see is closely related to the zero-noise case of \cref{delta} where the kernel in that case is a delta function. 
For discussion in the next section, we will specialize further to the truncated normal distribution, $s\sim t{\cal N}(0,\sigma)$ to maintain perturbations within the unit square thereby avoiding unbounded tails. The density function for the truncated normal distribution is given by:
\begin{equation}\label{tnormal}
    \nu(x;\mu,\sigma,a,b)=\frac{1}{\sigma} \frac{\phi(\frac{x-\mu}{\sigma}) }{\Phi(\frac{b-\mu}{\sigma})-\Phi(\frac{a-\mu}{\sigma})}, \mbox{ where, } \phi(z)=\frac{1}{\sqrt{2 \pi}} e^{-\frac{z^2}{2}}, \Phi(z)=\frac{1+\gls{erf}(\frac{z}{\sqrt{2}})}{2},
\end{equation}
and we choose, $a = 0,\ b = 1$, to define the subinterval, and $x-\mu=f(y)$.

\begin{figure}[htbp]
\centering
\includegraphics[width=0.4\textwidth]{ 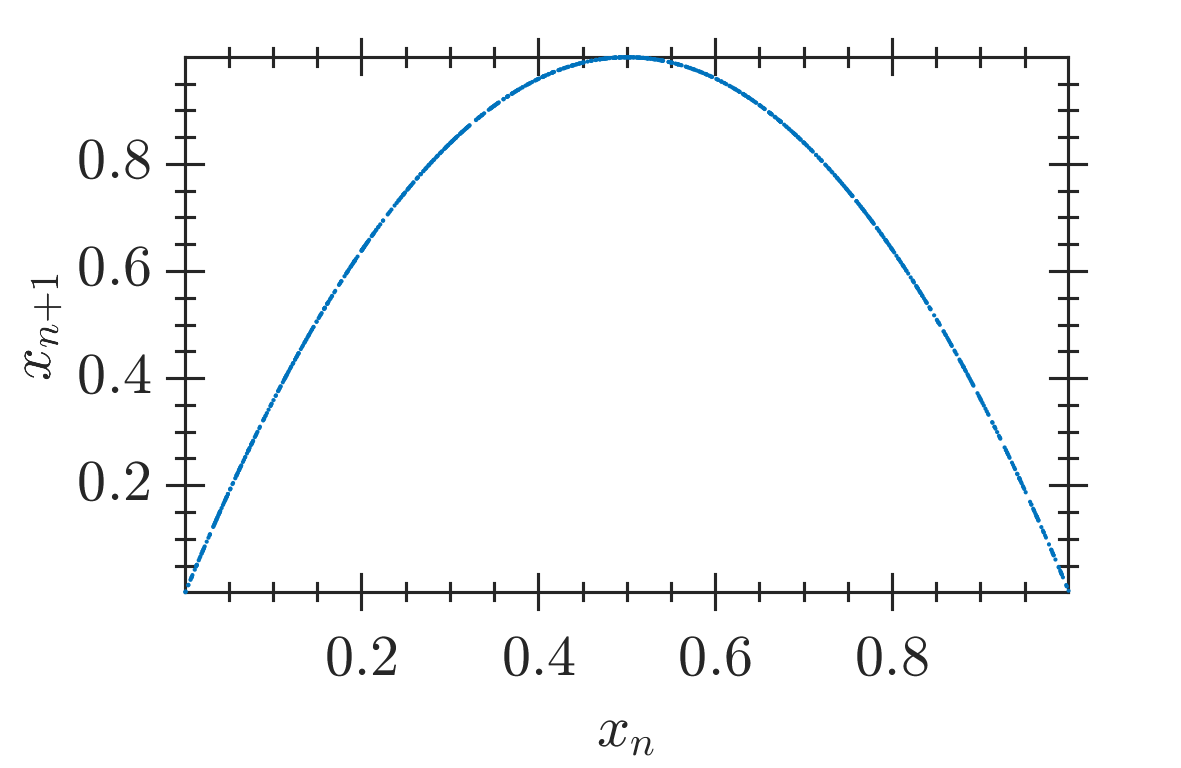}
\includegraphics[width=0.4\textwidth]{ 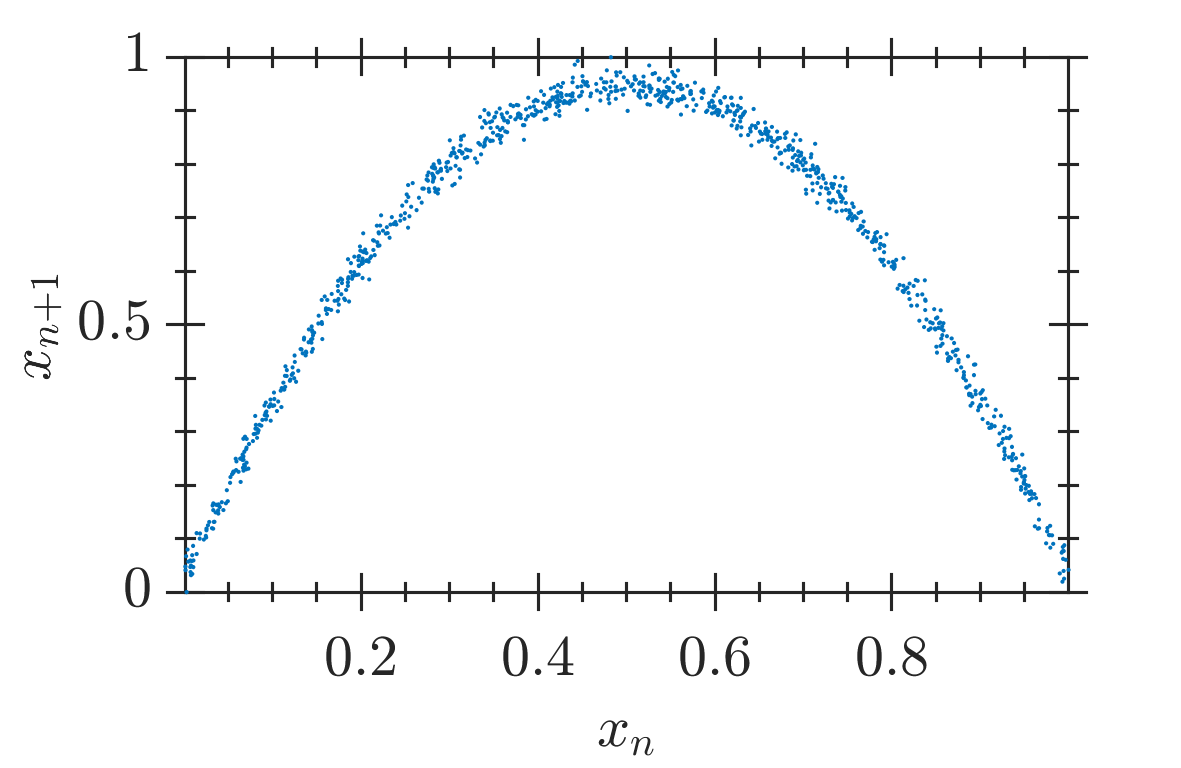}
\caption{ The figure presents data consisting of $N=1,000$ samples of $(x_n,x_{n+1})$ pairs, obtained from two different systems. The left panel shows a simulation obtained from the logistic map $x_{n+1}=f(x_n)=4 x_n (1-x_n)$, after an initial transient, so that the sample distribution closely approximates the invariant distribution $\rho_X(x)=\frac{1}{\pi\sqrt{x(1-x)}}$. On the right panel, the data is obtained from a noisy logistic map $x_{n+1}=f(x_n)=4 x_n (1-x_n)+s_n$, where $s_n$ is chosen from an independent and identically distributed (i.i.d.) truncated normal distribution with standard deviation $\sigma=0.02$.
}
\label{fig1}
\end{figure}

\begin{figure}[htbp]
\includegraphics[width=.3\textwidth]{ 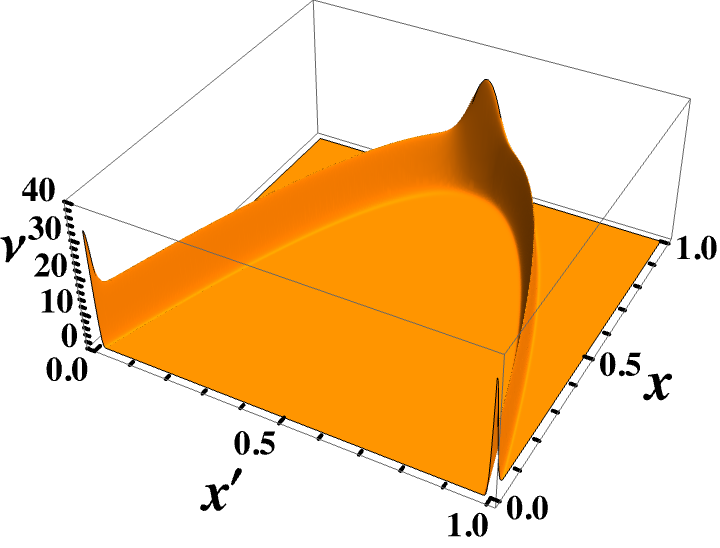}
\includegraphics[width=.3\textwidth]{ 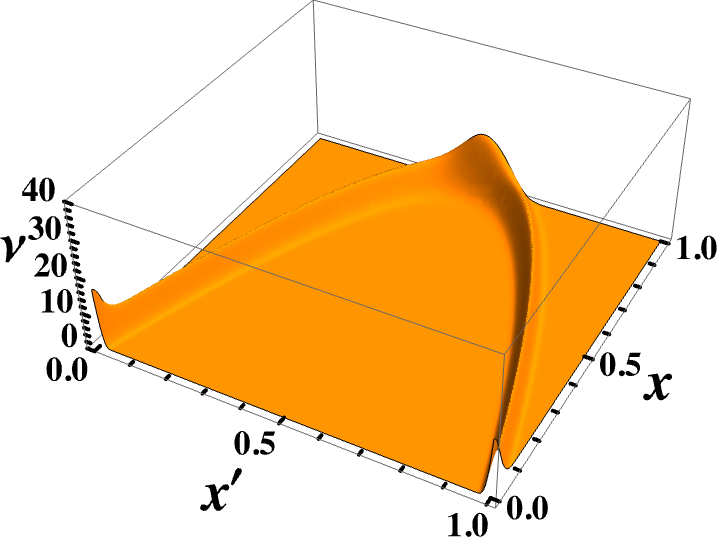}
\includegraphics[width=.3\textwidth]{ 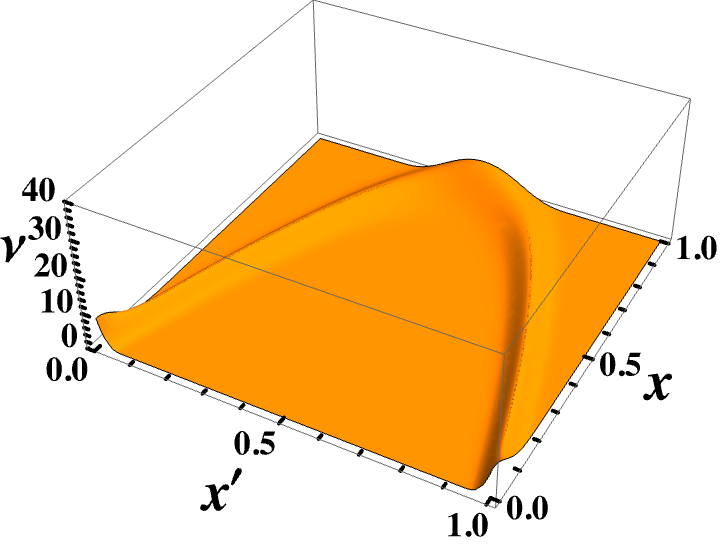}
\caption{This figure displays the kernel function ($\nu$) of the Frobenius-Perron operator for the truncated normal distribution, as described in \cref{tnormal}. The figure illustrates the kernel for three different standard deviations, namely $\sigma=0.025$, $\sigma=0.05$, and $\sigma=0.1$. The "bumps" observed in the distribution are a direct consequence of the requirement for a bounded domain, which ensures that the distribution remains a probability distribution with a unity integral. Additional insights can be obtained by referring to the sample data presented in \cref{fig1}.}
\label{figs}
\end{figure}

In \cite{bollt2013applied} the random sampling associated with \cref{est2} was interpreted as a Monte-Carlo integration estimate involved with projection onto basis functions as in \cref{est3}.
Now in the next sections, we will encode this same expression as a histogram-based density estimator of a Bayesian interpretation of the transfer operator.  This will open the door to considering a different kind of error analysis, as well as other estimators. 

A sample of data from the Logistic map, both unperturbed as well as with noise drawn from a truncated normal distribution, is shown in \cref{fig1,figs}. \cref{fig1} compares the sample data for the standard Logistic map and the perturbed version in the $(x_n, x_{n+1})$ plane. The data is obtained by taking $N=1,000$ samples of $(x_n, x_{n+1})$ pairs, with the former following approximately the invariant distribution $\rho_X(x)=\frac{1}{\pi\sqrt{x(1-x)}}$, while the latter system is obtained by adding noise to the standard Logistic map. \cref{figs} further discusses the kernel of the Frobenius-Perron operator for perturbed the Logistic map. The kernel is shown for different standard deviations, namely $\sigma=0.025$, $\sigma=0.05$, and $\sigma=0.1$, and reveals the distinctive "bumps" observed in the distribution. These bumps are a direct result of the bounded domain requirement, which ensures that the distribution remains a probability distribution with a unity integral, $\int_{\Omega} \rho(x)dx=1$. The insights obtained from this figure provide a better understanding of the effect of noise on the dynamics of the system.  

\begin{figure}[htbp]
\centering
\includegraphics[width=0.32\textwidth]{ 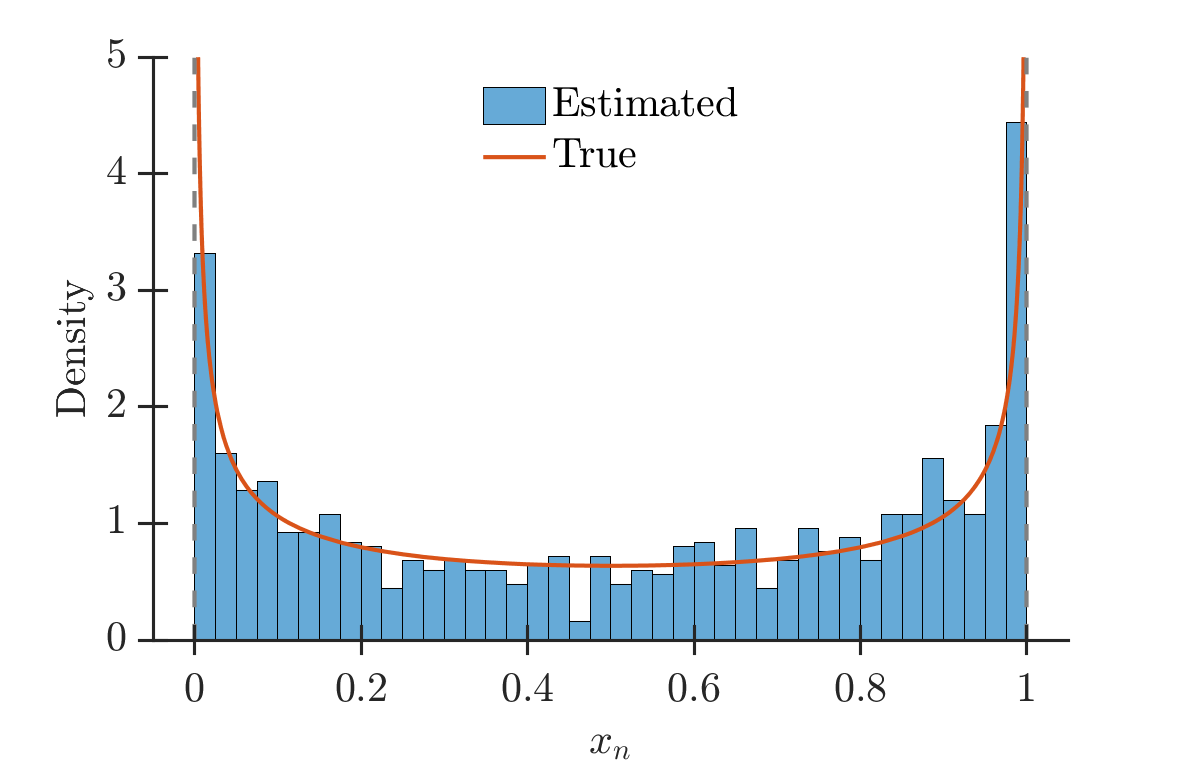}
\includegraphics[width=0.32\textwidth]{ 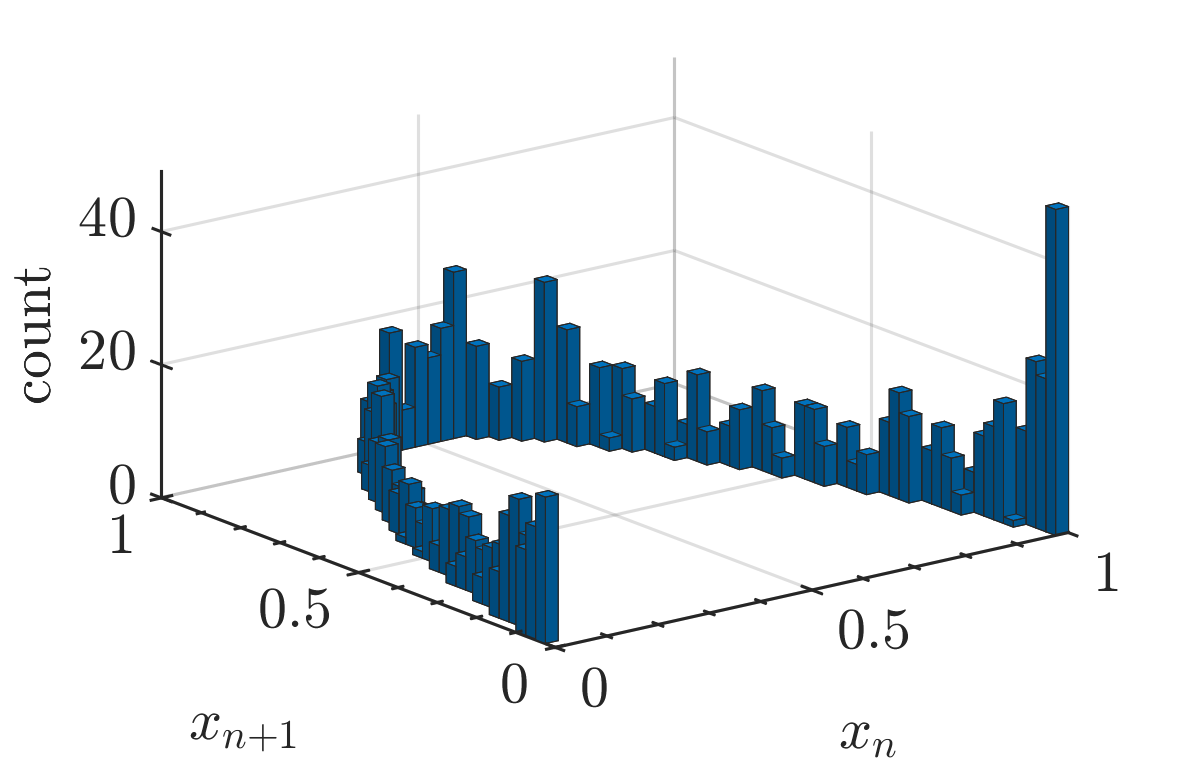}
\includegraphics[width=0.32\textwidth]{ 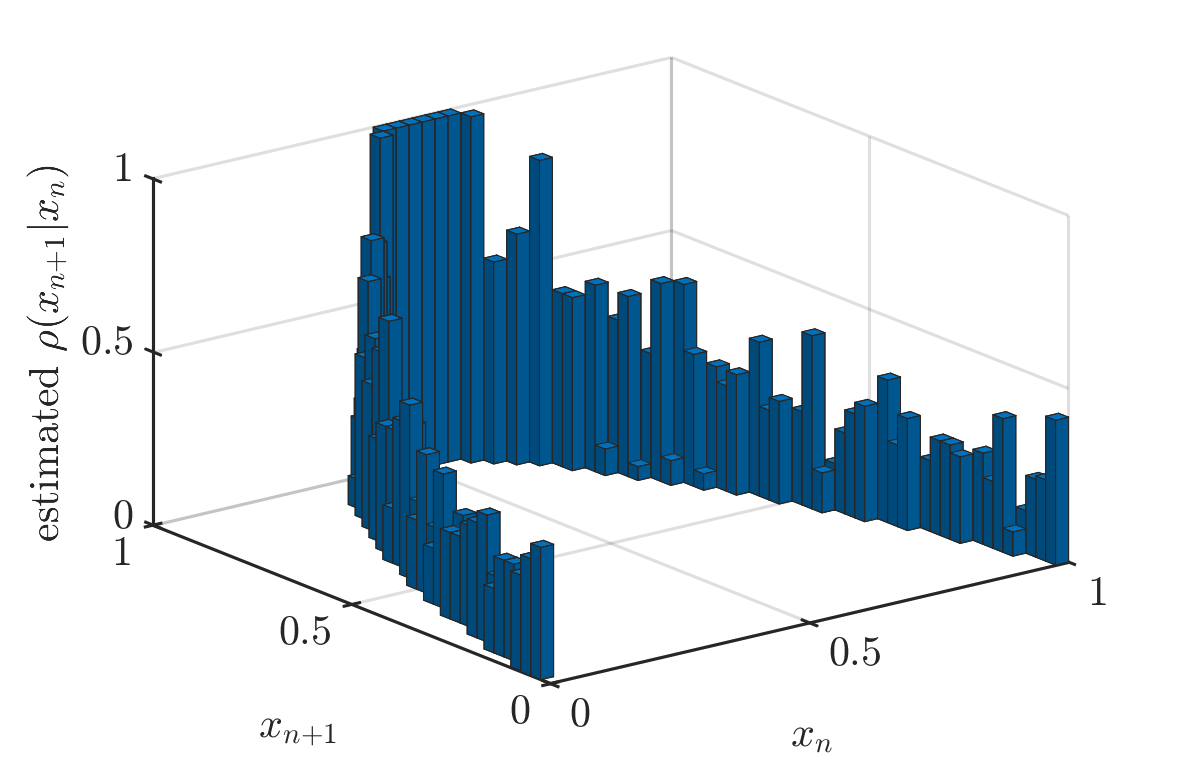}
\includegraphics[width=0.32\textwidth]{ 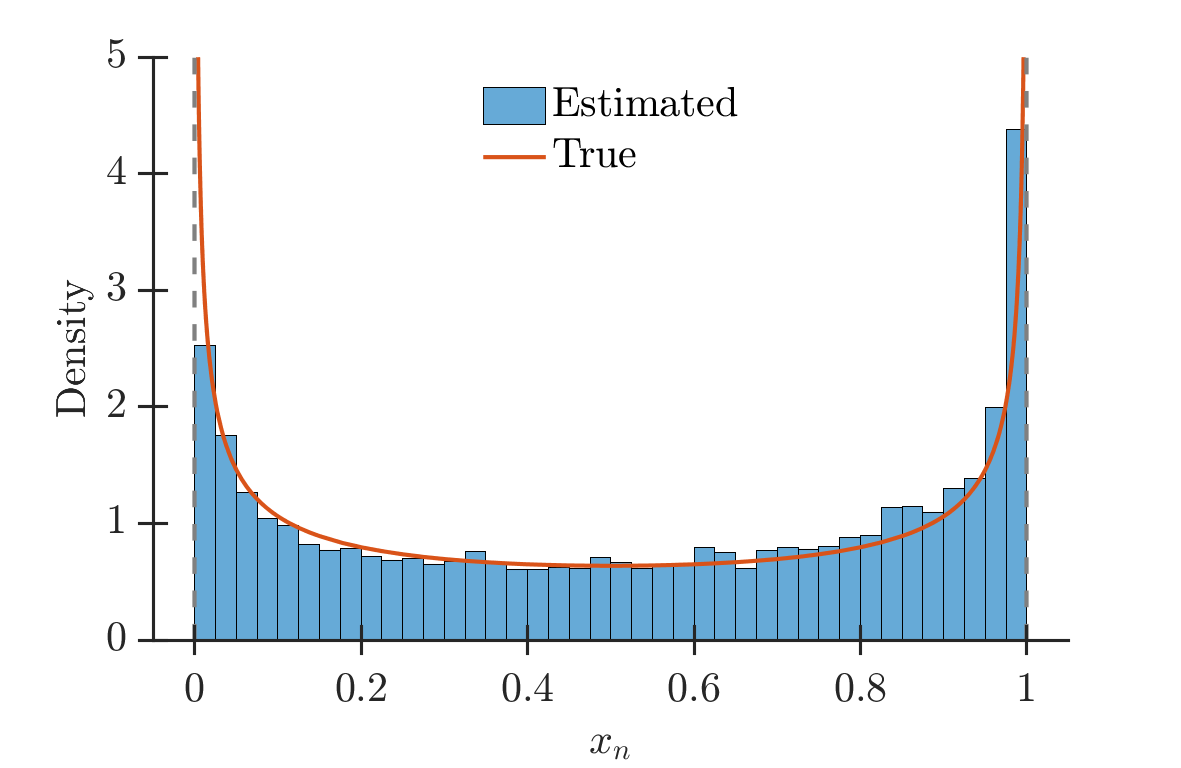}
\includegraphics[width=0.32\textwidth]{ 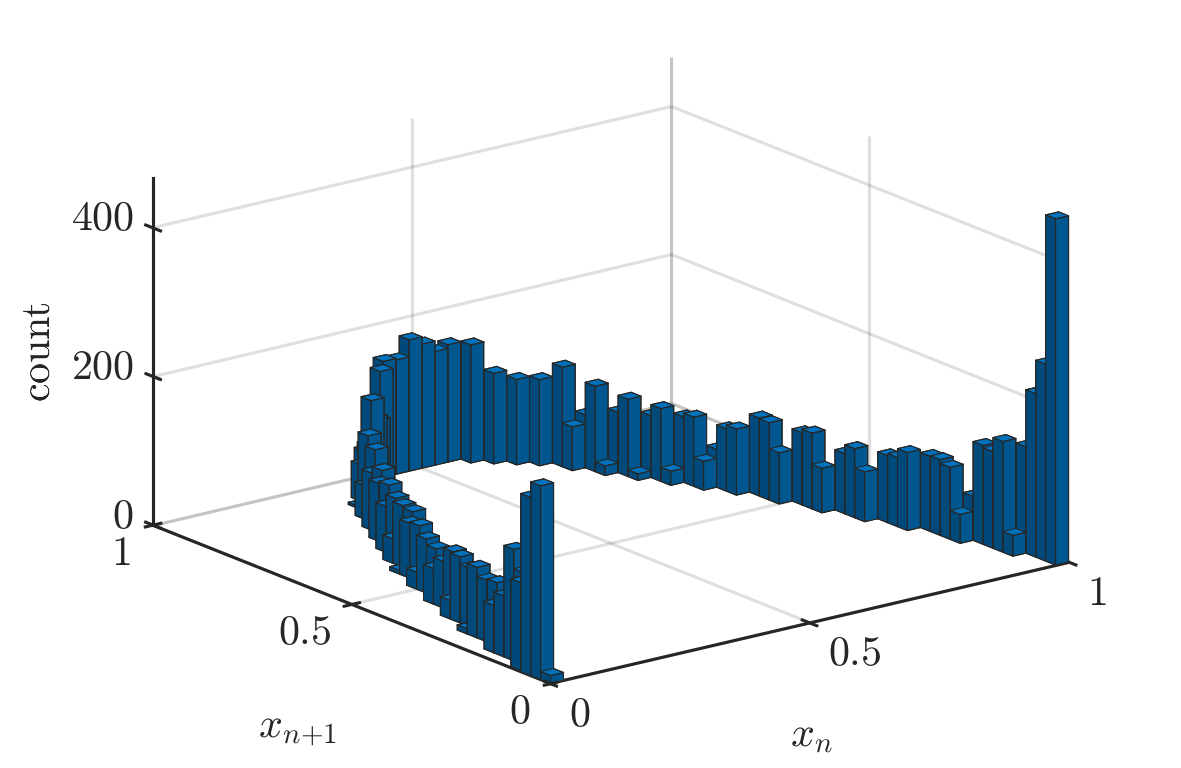}
\includegraphics[width=0.32\textwidth]{ 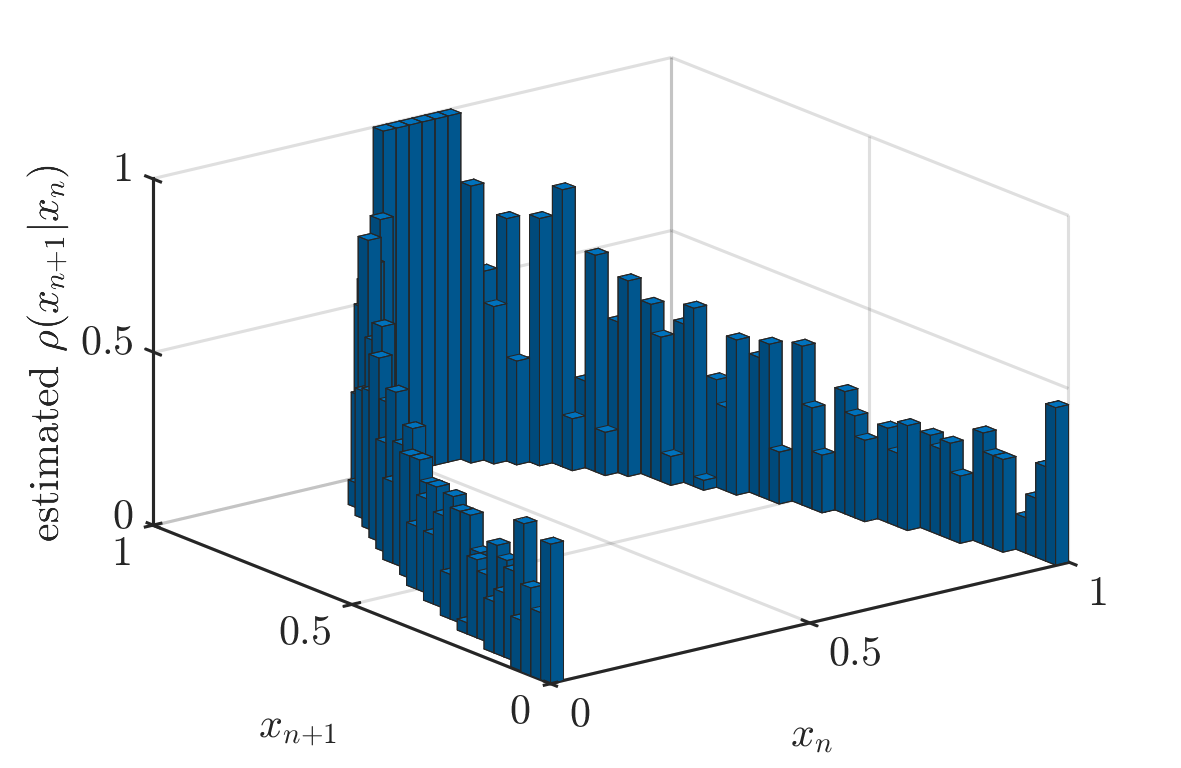}
\caption{ The figure shows the \gls{hde} for data $(x_n,x_{n+1})$ from sample orbits of size $N=1,000$ (top row) and $N=10,000$ (bottom row) using the logistic map (noise-free).  The bin-width used for estimation is $1/K$ with $K=40$, and the joint distribution estimate employs a grid of $40 \times 40$ cells. The bottom row, which represents higher sample orbits, provides smoother and more accurate estimates of the true distributions with reduced variability. Comparing the estimated marginal distributions to the true distribution (left figures) allows for insights into the effectiveness of the estimation techniques employed in this study. The relatively higher densities observed around points $(0,0)$, $(0,1)$, and $(0.5,1)$ in the middle and right figures indicate changes in data concentration from $x_n$ to $x_{n+1}$.}
\label{fig2}
\end{figure}

\begin{figure}[htbp]
\centering
\includegraphics[width=0.32\textwidth]{ 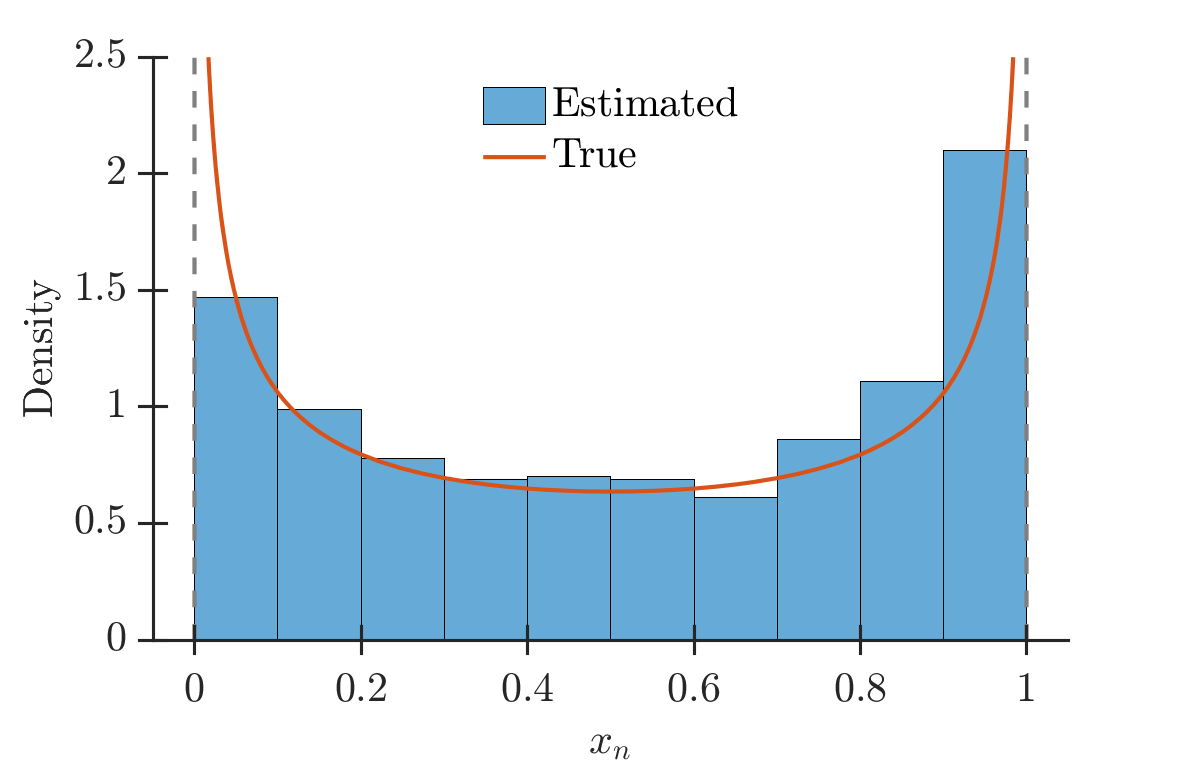}
\includegraphics[width=0.32\textwidth]{ 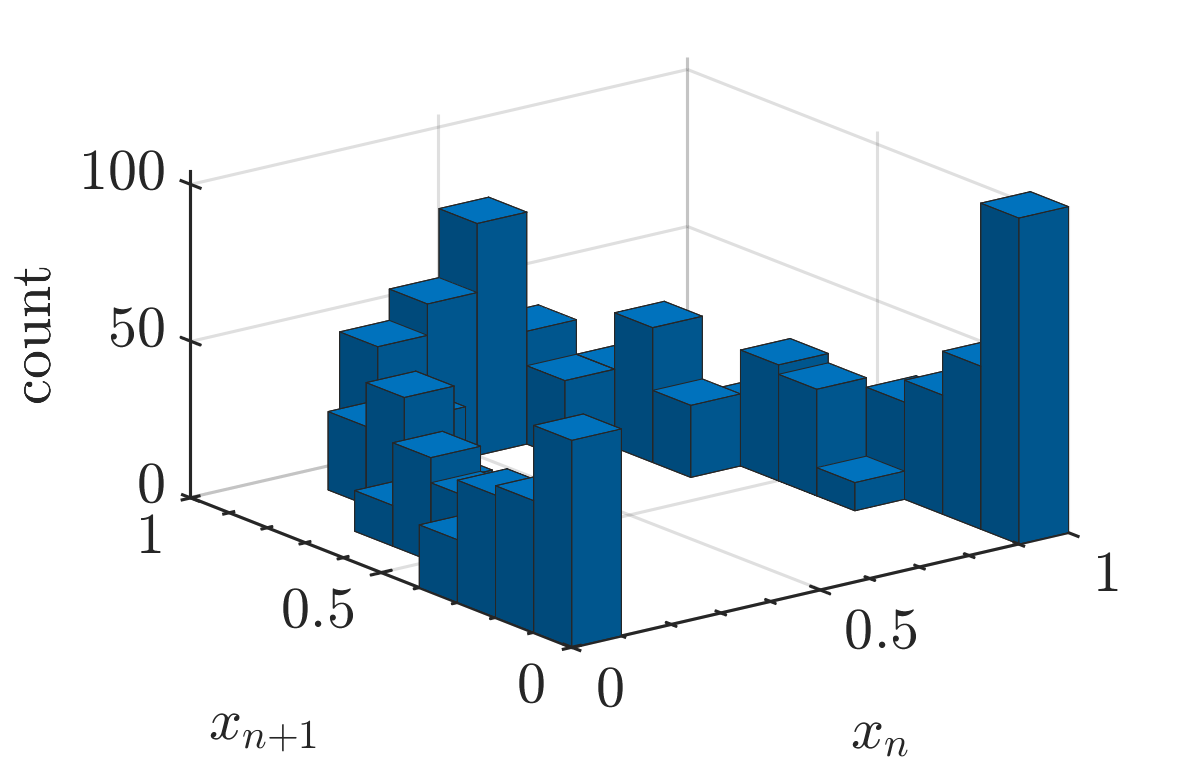}
\includegraphics[width=0.32\textwidth]{ 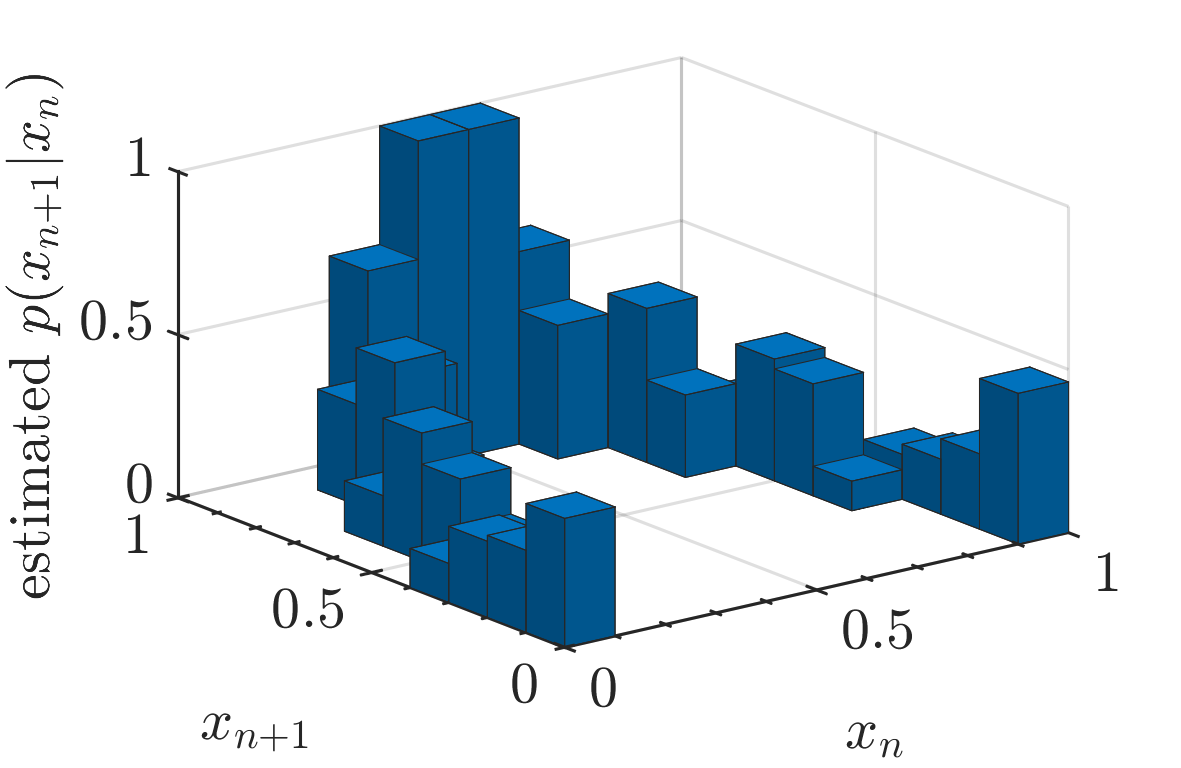}
\caption{ The data in this figure is similar to that presented in \cref{fig2}, but it uses \gls{hde} with a coarser resolution. The estimation employs $K=10$ and $10 \times 10$ cells for marginal, joint, and conditional densities, resulting in a wider bandwidth that reduces variability but increases bias.} 
\label{fig4}
\end{figure}

\begin{figure}[htbp]
\centering
\includegraphics[width=0.32\textwidth]{ 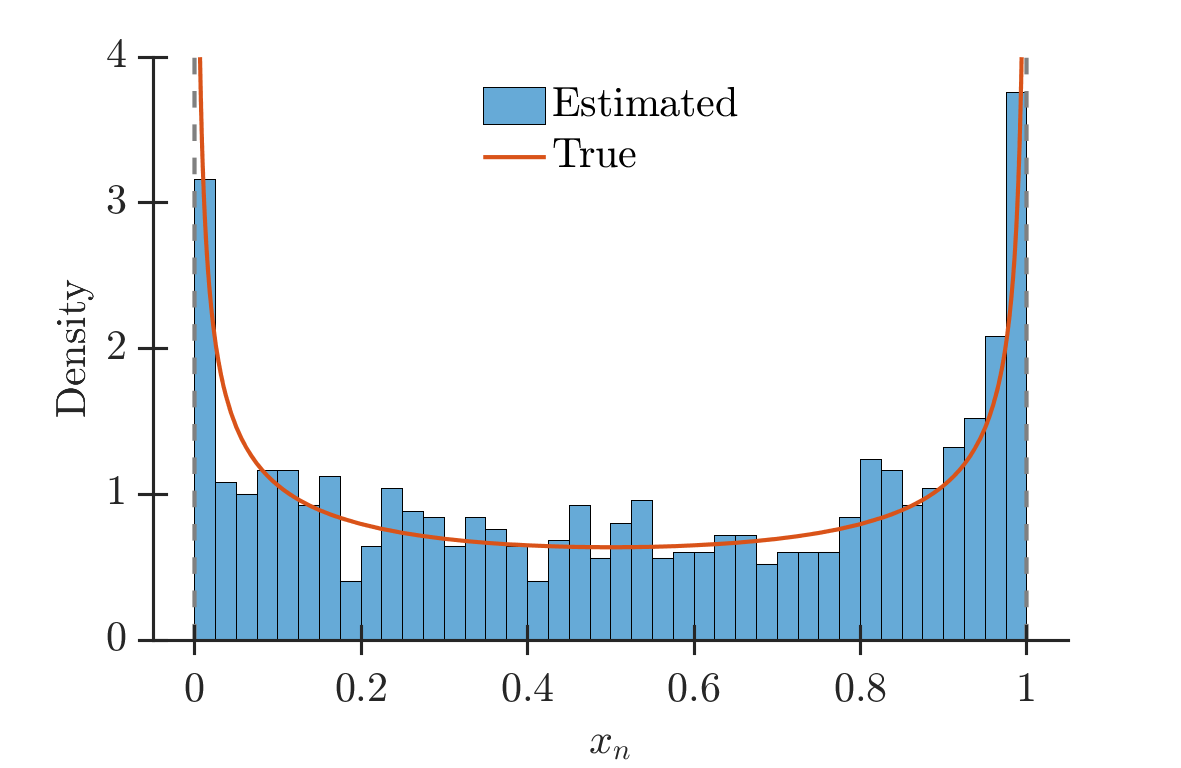}
\includegraphics[width=0.32\textwidth]{ 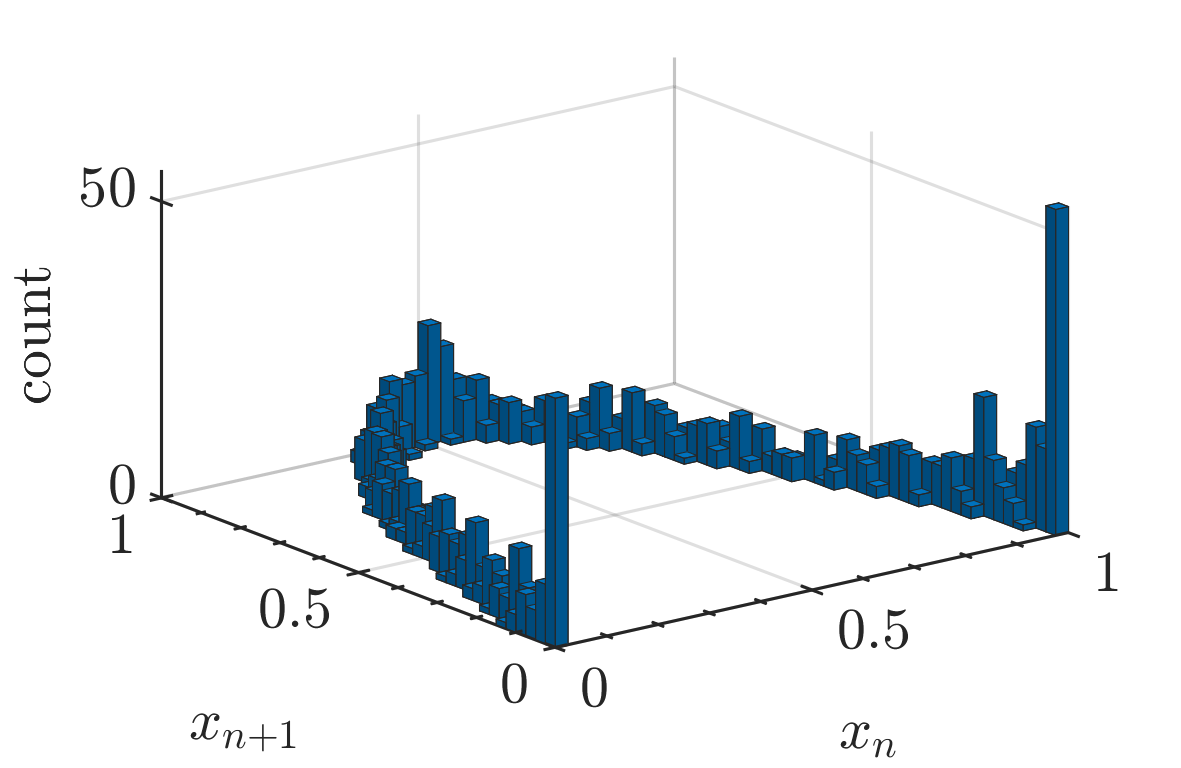}
\includegraphics[width=0.32\textwidth]{ 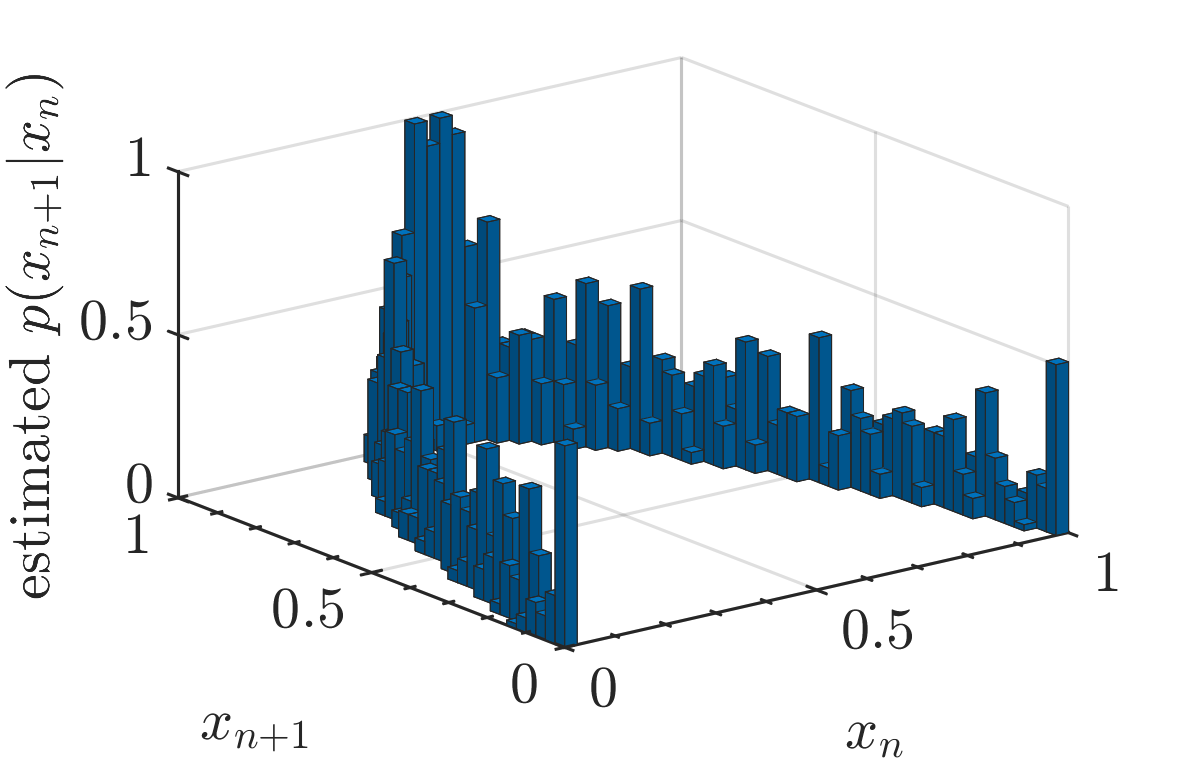}
\caption{ The figure showcases the \gls{hde}, similar to \cref{fig4} using data from a random Logistic map orbit with truncated normal distribution noise ($\sigma=0.02$) as depicted in  \cref{fig1}. These plots exhibit similar characteristics of variability versus bias (smoothing) as observed in the noiseless scenario of \cref{fig2}(Top Row), even with the same level of smoothing, despite the differences in the true underlying distribution. However, it is noteworthy that the marginal distribution estimate of the invariant distribution appears significantly less smooth.}
\label{fig5}
\end{figure}

Estimating the density of the logistic map data $(x_n, x_{n+1})$ using the histogram estimation method, as shown in \cref{fig2,fig4,fig5}, provides a starting point for discussing density estimation theory while reinterpreting the estimation of the Frobenius-Perron operator using the Ulam method as a histogram density estimation method. In these figures, we estimate the distributions $\rho_X(x)$ (left), $\rho_{X'X} (x',x)$ (middle), and $\rho_{X'|X}(x'|x)$ (right). These figures offer insights into various aspects of estimation. For instance, Figure \cref{fig2} showcases these estimates for a sample orbit of size $N=1,000$ (top row) and $N=10,000$ (bottom row) for the logistic map. The estimation is performed using a bin-width of $1/K$ with $K=40$, and a grid of $40 \times 40$ cells is employed for the joint distribution estimate. The rightmost estimate, $\rho_{X'|X}(x'|x)$, is obtained using the methods described in \cref{est1,est2,est3,est4,eq:UG}. The larger sample orbit in the bottom row provides smoother and more accurate estimates of the true distributions with reduced variability.

Additionally, in \cref{fig4}, estimation is performed with $K=10$ and $10 \times 10$ cells for the marginal, joint, and conditional distributions. This results in less variance at the expense of increased bias due to the wider bandwidth (see, \cref{Sec:TheoryDen}). Furthermore, \cref{fig5} showcases the histogram density estimates for data from a random logistic map orbit with truncated normal distribution noise ($\sigma=0.02$). These estimates exhibit similar variability versus bias characteristics to the noiseless scenario, even with the presence of noise. However, it is worth noting that the marginal distribution estimate of the invariant distribution appears significantly less smooth in this case.

Overall, \cref{fig2,fig4,fig5} provide valuable insights into the estimated distributions $\rho_X(x)$, $\rho_{X'X}(x',x)$, and $\rho_{X'|X}(x'|x)$, demonstrating the effects of sample size ($N$), bin size ($K$), and noise on the accuracy and smoothness of the estimates. Additionally, $\rho_{X'|X}(x'|x)$ (rightmost figures) show Ulam-Galerkin estimates of the stochastic matrices (equation \cref{est3}) interpreted as histogram estimators, which will be further discussed in the subsequent section on corresponding Bayesian estimators.

\section{Bayesian Interpretation of the Transfer Operator}\label{Sec:BayInter}

The Frobenius-Perron operator can be interpreted in a Bayesian context in the following manner. Let us denote the output-input pair of the function $f$ by $(x',x)$, where $x'=f(x)$, and consider these pairs as samples of the random variables $X$ and $X'$. Note that $X'$ is a transformed version of the random variable $X$ under the function $f$. A statement of conditional and compound densities leads to an interpretation of the Frobenius-Perron operator as a Bayes update.  Reviewing, the joint density $\rho_{X'X}(x',x)$, of random variables $X'$ and $X$ marginalizes to,
\begin{equation}
    \rho_{X'}(x')=\sum_{x:x'=f(x)} \rho_{X'X}(x',x)=\sum_{x:x'=f(x)} \frac{\rho_X(x)}{|Df(x)|},
\end{equation}
in terms of the summation of all pre-images of $x'$.  Notice that the middle term is written as a marginalization across $x$ of all those $x$ that lead to $x'$.  This Frobenius-Perron operator, as usual, maps densities of ensembles under the action of the map $f$. Compared to the defining statement of a conditional density in terms of a joint density,
\begin{equation}
    \rho_{X'X}(x',x)=\rho_{X'|X}(x'|x)\rho_X(x).
\end{equation}
We reinterpret, in the noiseless case, 
\begin{equation}
    \rho_{X'|X}(x'|x)=\frac{1}{|Df(x)|} \boldsymbol{\delta}(x'-f(x)).
\end{equation}
Here, $\boldsymbol{\delta}(x)$ is interpreted as the Kronecker delta function, but if the summation is an integral, $\int_{\Omega} \rho_{X'X}(x',x) dx$, then it can be interpreted as a Dirac delta function. 

In the language of Bayesian uncertainty propagation, $\rho_{X'|X}(x'|x)$ describes a likelihood function, interpreting future states $x'$ as data and past states $x$ as parameters, by the standard Bayes phrasing, 
\begin{equation}
\rho(\Theta|\mbox{data})\propto \rho(\mbox{data}|\Theta)\times \rho(\Theta),
\end{equation}
for parameter $\Theta$, or simply by standard names of the terms, 
\begin{equation}
\mbox{posterior}\propto\mbox{likelihood}\times \mbox{prior}.
\end{equation}
From a Bayesian perspective, we can interpret the elements of the matrix approximation $\gls{eFP}$ in a manner analogous to Equation \ref{est1}. Specifically, the elements $\gls{eFP}_{i,j}$ can be regarded as a matrix of likelihood functions, given by:

\begin{equation}\label{est4}
\gls{eFP}_{i,j}=P(x'\in B_j|x\in B_i) = \frac{P(x\in B_i, x'\in B_j)}{P(x\in B_i)} = \frac{m(B_i\cap f^{-1}(B_j))}{m(B_i)}=\frac{\int_{B_i\cap f^{-1}(B_j)}dm(x)}{\int_{B_i}dm(x)}.
\end{equation}

Here, $dm(x)$ represents the differential of the probability measure $m$, $m(B_i)$ represents the prior probability of $B_i$, and $m(B_i\cap f^{-1}(B_j))$ represents the joint probability of $B_i$ and the pre-image of $B_j$ under the transformation $f$. By comparing this interpretation to that of Equation \ref{est1}, we can see that the matrix $\gls{eFP}$ can be thought of as a set of conditional probabilities that quantify the likelihood of transitioning from one bin to another under the transformation $f$.
Furthermore, the standard Ulam estimator, \cref{est2}, can be taken as  a histogram method to estimate, the joint and marginal probabilities, $\rho_{X'X}$ and $\rho_X$. This estimate is given by occupancy counts in the related boxes, $B_i$ and $B_j$ with,
\begin{equation}\label{eq:UG}
    \begin{aligned}
        P(x\in B_i, x'\in B_j)&\sim& &\# x_n\in B_i, x_{n+1}\in B_j, \mbox{ and, }  \\ 
        P(x\in B_i)&\sim& &\# x_n\in B_i.
    \end{aligned}
\end{equation}

The conditional follows by division, to the estimator of the matrix $\gls{eFP}_{i,j}$ describing the likelihood function.  In these terms, we are positioned to describe the statistical error of expressions such as \cref{est2} for the matrix $\gls{eFP}_{i,j}$ estimator of the Frobenius-Perron operator, by the theory of density estimators, for $\rho_{X'X}(x',x)$ and $\rho_X(x)$ respectively.  First, in the next section, we will discuss this histogram estimator, and then in the following sections, we will consider other estimators, notably Kernel Density Estimation (\gls{kde}).

\begin{figure}[htbp]
\includegraphics[width=0.45\textwidth]{ 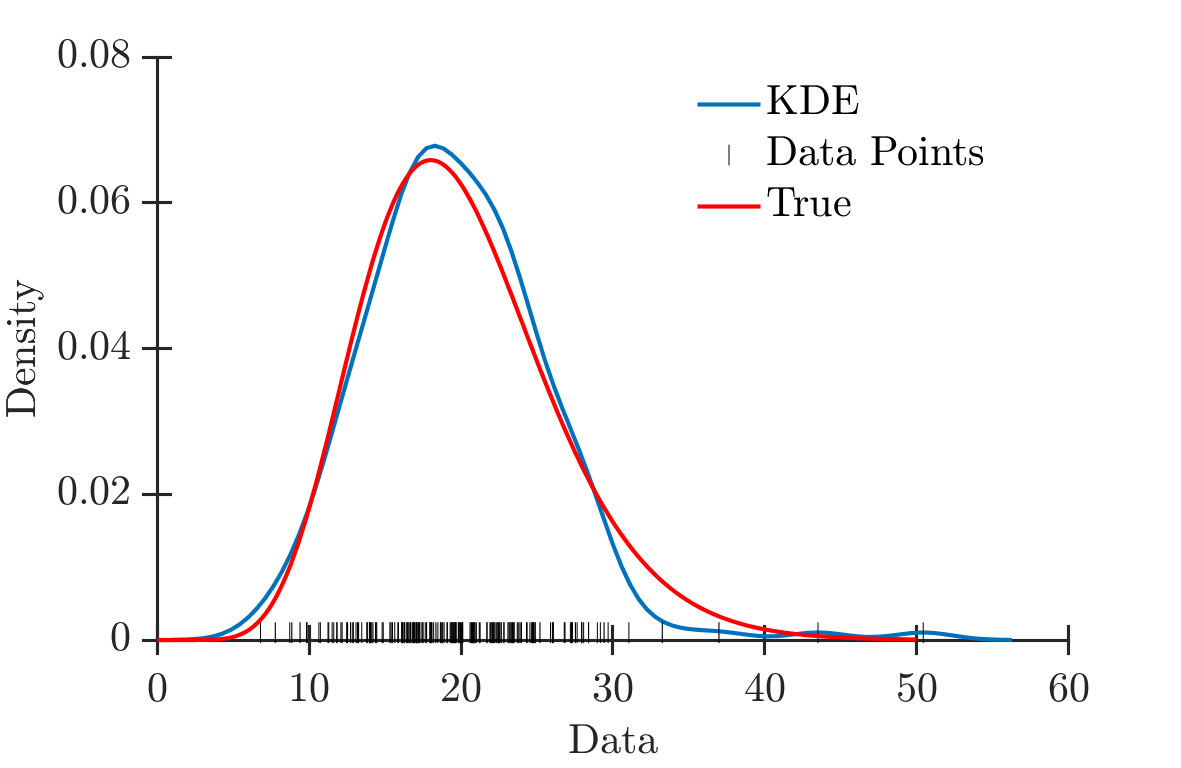}
\includegraphics[width=0.45\textwidth]{ 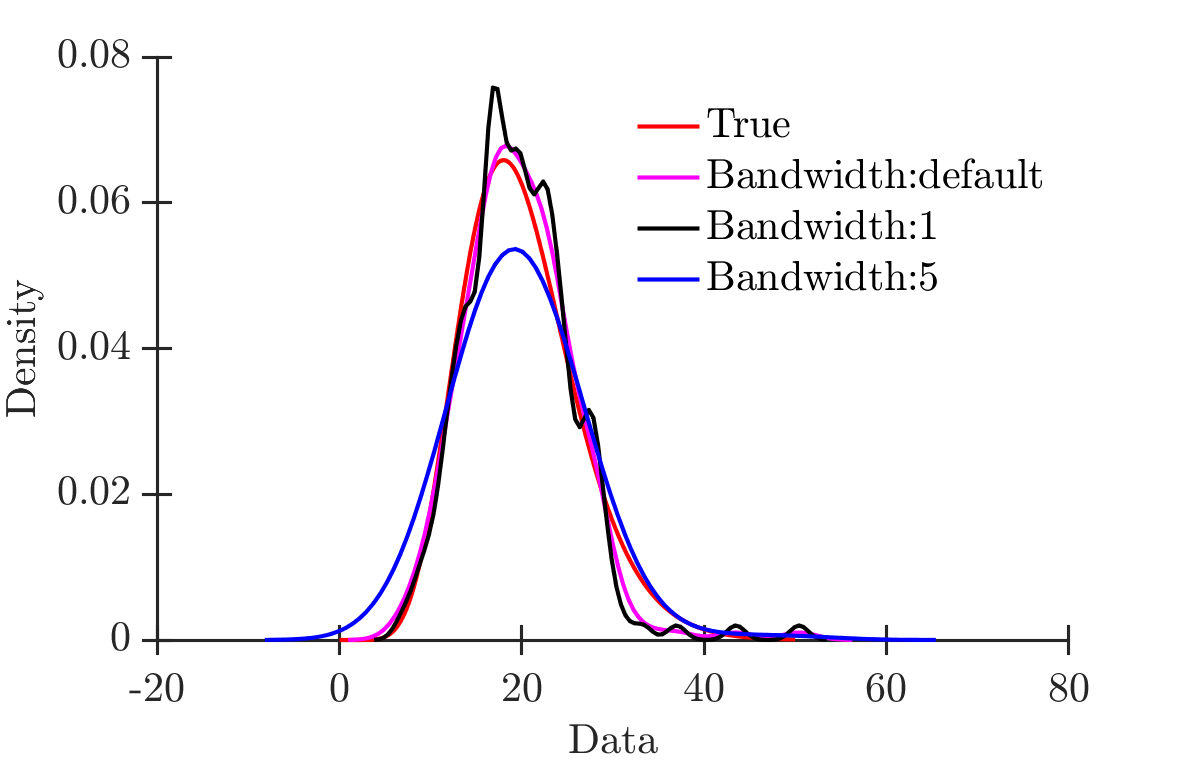}\\
\includegraphics[width=0.45\textwidth]{ 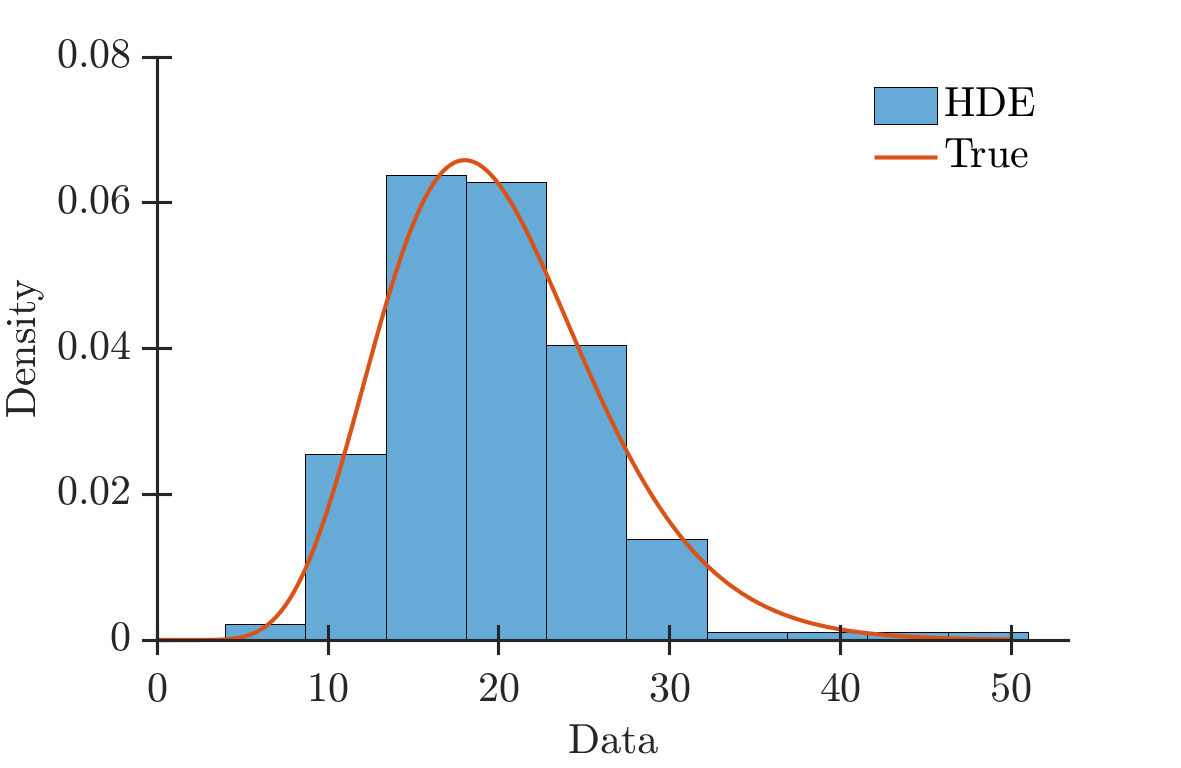}
\includegraphics[width=0.45\textwidth]{ 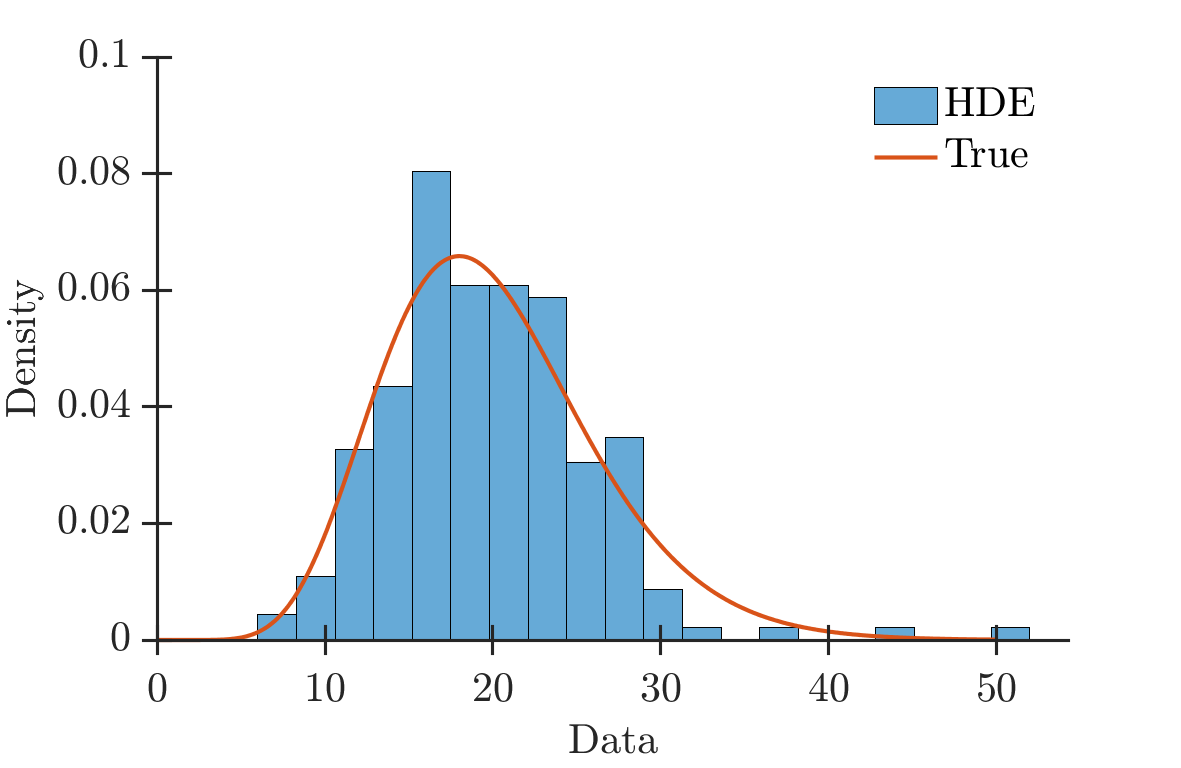}
\caption{The figures in the top row present \gls{kde} estimation using the Gaussian kernel for a specific data set generated from the Chi-squared distribution with 20 degrees of freedom (the true density function is depicted as the red curve in each plot). The top right figure showcases the comparison of estimations with different kernel bandwidths. The bottom row demonstrates density estimation for the same data set using the histogram estimation method. Furthermore, the bottom left and right figures compare the estimations with different bin sizes ($K=10$ and $K=20$ respectively).}
\label{fig6}
\end{figure}

\section{Theory of Density Estimation }\label{Sec:TheoryDen}

As we have argued in the previous section, the problem of estimation of a Ulam-Galerkin estimator of the Frobenius-Perron operator is equivalent to the Bayes computation of the conditional density $\rho_{X'|X}(x'|x) $, derived by histogram estimators of the joint and marginal densities,  $\rho_{X'X}(x',x),\ $ $\rho_X(x),$ respectively.  Therefore in this section, we review what is a classical theory from the statistics of density estimation, found in many excellent textbooks, such as \cite{silverman_1999,scott_2015}.  First, we will review some details of histogram estimators, considering the central issues \gls{bias}, variance, and choice of bandwidth.  Then in the subsequent subsection, we will re-cast the problem as one of the \gls{kde}, for which those same three issues, \gls{bias}, variance, and bandwidth, suggest some advantages relative to histogram estimation.

\cref{fig6} presents a motivating example for the subsequent discussion, utilizing data derived from the chi-square distribution with 20 degrees of freedom. In this article, we employ numerical calculations using MATLAB's built-in functions, specifically ``ksdensity" for \gls{kde} and ``histogram" for \gls{hde}. The default bandwidth of the ``ksdensity" function is optimized for estimating assuming a Gaussian Kernel\cite{MATLAB:2022}. 

The top row of \cref{fig6} compares the \gls{kde} of the data with the true probability density function (represented by the red curve in the plots). The top-left figure illustrates the estimated density using the default bandwidth from "ksdensity", while the top-right figure showcases the variations in the estimated density with different bandwidth selections for the \gls{kde} method. The bottom row of \cref{fig6} contrasts the estimated density using \gls{hde} with a bin size of $K=10$ (left) versus $K=20$ (right). It is important to note that reducing the bandwidth or increasing the bin size does not necessarily result in more accurate estimation. There exists an optimal bandwidth or bin size for these methods, dependent on the number of data points and the characteristics of the density function. This section delves into the mathematical intricacies of these estimation methods by investigating the mean squared error (\gls{mse}).

For further discussion, we consider a general random variable $X$ distributed according to $\rho_X(x)$. However, to simplify the analysis, this section assumes a unit interval, $x \in [0,1]$, which also serves as the support of $\rho_X(x)$.  Likewise, assume $(x',x) \in [0,1]\times [0,1]$, and $\rho_{X'X}(x',x)$ has support in the unit square.  The standard theory of density estimation also assumes a smooth density function\cite{silverman_1999,scott_2015,fan2018local}, $|\rho_X'(x)|\leq C_1$ and $|D\rho_{X'X}|\leq C_2$, for constants of uniform bound, $C1, C_2\geq 0$.   Note however, this is already a problem regarding perhaps the most popular example for pedagogical study, the invariant density of the logistic map is $\rho_x(x)=\frac{1}{\pi\sqrt{x(1-x)}}$, is unbounded and has an unbounded derivative, $\rho'_X(x)=\frac{2x-1}{2\pi \sqrt{x(1-x)}^3}$. Likewise, Markov and semi-Markov maps will not generally have smooth densities as necessary conditions for the application of the estimation theory with bounds. However, like the density of the true Logistic map, these  too have smooth densities when the noise of the mapping is used as well as its corresponding transfer operator.  Nonetheless, many have made a practice of presenting the invariant distribution as estimated from sample orbits, \cref{fig2} (Left). Additionally, there exist numerical techniques for selecting an appropriate bandwidth in order to mitigate the challenges associated with boundary effects \cite{sheather1991reliable}.

The key issue in any estimator is  accuracy versus the amount of data available. Generally, we want an analysis of \gls{mse} of the estimator, which requires both \gls{bias} and variance, since $\gls{mse}=\gls{bias}^2+\gls{var}$.

\subsection{Theory of Density Estimation Using Histograms}

Here we review density estimation, closely following \cite[Section 3.2]{scott_2015}, first in one dimension, and then the multivariate scenario.

Considering a unit interval, $[0,1]$, it may be divided into $K$ cells (bins), 
\begin{equation}
    {\cal B}=\{B_i\}_{i=1}^K, B_i=\Big[\frac{i-1}{K},\frac{i}{K}\Big), i=1,2,..,K,
\end{equation}
which is a uniform topological partition, meaning interiors are mutually disjoint and the union closure covers (both closed under taking unions and closures and also forms a cover). Similarly, a multivariate histogram is a topological partition into bins, usually rectangles (but other shapes, especially tessellations are not uncommon).  Otherwise, we are continuing with the discussion of the single variate estimator. Given a sample $\{x_n\}_{n=1}^N$ and a set of bins $\{B_j\}_{j=1}^K$, the \gls{hde} for the random variable $X$, $\overline{\rho_{N,K}}(x_i)$, which approximates $\rho(x_i)$, for any $x_i \in B_i$, is given by 
\begin{equation}\label{histest}
    \overline{\rho_{N,K}}(x_i) = \frac{\# x_n \in B_i}{N} \times \frac{1}{m(B_i)} = \frac{K}{N}\sum_{j=1}^N \xi_{B_i}(x_j),
\end{equation}
 where $\# x_n \in B_i$ is the number of observations in bin $B_i$, $N$ is the total number of observations, $m(B_i)$ is the measure of bin $B_i$, and $\xi_{B_i}(x_j)$ is the indicator function as defined in \cref{eq:indicFac}.

The key part of density estimators is the analysis of the \gls{bias} of the estimator, \cite{scott_2015}, continuing with $\overline{\rho_{N,K}}(x_i) $ for a point $x_i\in B_i$ that need not be assumed to be one of the data points $x_j$.  Consider the probability that the sample $x_j\in B_i$, $P(x_j\in B_j)$,
\begin{equation}
    {\mathbb E}(\overline{\rho_{N,K}}(x_i))={\mathbb E}(K P(x_j\in B_j))=K\int_{\frac{i-1}{K}}^{\frac{i}{K}} \rho_X(s)ds=\rho_X(\tilde{x}), \mbox{ for } \tilde{x}\in B_j,
\end{equation}
by mean value theorem and the fundamental theorem of calculus.  Therefore, the \gls{bias} of the estimator is,
\begin{equation}\label{eq:histBias}
    \gls{bias}_{hist}[\overline{\rho_{N,K}}(x)]={\mathbb E}(\overline{\rho_{N,K}}(x)-\rho_X(x))=\rho_X(\tilde{x}) -\rho_X(x)\leq |\rho_X'(\hat{x})| |\tilde{x}-x| \leq \frac{C_1}{K}.
\end{equation}
The first inequality follows again by the mean value theorem, this time for a value $\hat{x}\in (x,\tilde{x})$ (or perhaps the opposite order), and a by-product of absolute values, and the fact that $\tilde{x},\hat{x}\in B_i$.  Furthermore, it should be noted that the issue of bias in \gls{hde} (see \cref{eq:histBias}) involves a trade-off  between limiting the derivative of the density function $\rho'_X\leq C_1$ and selecting an appropriate number of bins $K$ to obtain a desirable level of accuracy.

The variance can be computed with \cref{histest},
\begin{eqnarray}
    \gls{var}_{hist}\Big(\overline{\rho_{N,K}(x)}\Big)&=&K^2 \gls{var}\Bigg(\frac{K}{N}\sum_{j=1}^N \xi_{B_i}(x_j)\Bigg) \nonumber \\
    &=& \frac{K^2 P(x_j\in B_i)(1-P(x_j \in B_i))}{N}=
\frac{K^2\Big( \frac{\rho_X(\tilde{x})}{K} \Big)\Big(1- \frac{\rho_X(\tilde{x})}{K}  \Big)}
{N} \nonumber \nonumber \\
&=& \frac{K \rho_X(\tilde{x}) +  \rho_X^2(\tilde{x})}{N}.
\end{eqnarray}
Variance is a question of balancing the number of bins $K$ versus the  data count $N$, but relative to the unknown density $\rho_X$.

Therefore the mean square error for the density estimation $\rho_{N,K}(x)$ at an arbitrary point $x\in [0,1]$ follows, 
\begin{equation}\label{msehist}
    \gls{mse}_{hist}(\overline{\rho_{N,K}}(x))=\gls{bias}_{hist}^2(\overline{\rho_{N,K}}(x))+\gls{var}_{hist}(\overline{\rho_{N,K}}(x))\leq \frac{C_1^2}{K^2}+
    \frac{K \rho_X(\tilde{x}) +  \rho_X^2(\tilde{x})}{N}.
\end{equation}

To interpret, when a fixed data set of size $N$ is given, from an unknown distribution $\rho_X$, we can only choose $K$ and this choice is called bandwidth selection.  From \cref{msehist}, large $K$ (more bins) yields decreased \gls{bias} (the first term), but the variance (the second term) will tend to be large. This demonstrates the balancing struggle between \gls{bias} and variance in choosing the number of bins, and the bandwidth.  Figs.~\ref{fig2}-\ref{fig5} demonstrate this bandwidth selection balancing act.  Again, we reiterate that formally the analysis requires $C_1\geq 0$ be bounded whereas the derivative of the invariant density of the logistic map is not of bounded derivative type in $[0,1]$, still many estimates exist in the literature (\cite{Peter1995, bollt2000controlling, nie_coca_2018} etc.), including ourselves, which we now call a ``typical sin."  An argument that it may not be fatal for practical problems is the fact that in real-world dynamical systems, there is always noise, which has the effect of smoothing (e.g., noise sampled from a smooth distribution serves as a mollifier that can bring even a singular distribution ``blurred" into a $C^\infty$ distribution) or rather producing invariant densities that are smooth after all \cite{petersen1989ergodic,robinson1998dynamical,dajani2002ergodic,becker1989dynamical,tu2012dynamical}.  See \cref{fig2} for histogram density estimations for \cref{fig1} (Right) randomly perturbed logistic map data.

Note that analysis of \gls{mse} in the theory of multivariate histogram estimators is similar in methodology, to which we refer to \cite{scott_2015}.  The important point to this stage of the paper is that the famous Ulam-Galerkin estimation of the transfer operators by the formula \cref{est2} amounts to problems of density estimations of the marginal and joint distributions $\rho_X(x)$ and $\rho_{X'X}(x',x)$ leading to the estimation of the conditional distribution $\rho_{X'|X}(x'|x)$.  This said, we can now state that this is not a description of the Ulam problem (vs the Ulam-Galerkin estimation). The Ulam problem describes these estimates in terms of stochastic matrices, each with dominant eigenvector describing the invariant state of the corresponding Markov chain, and that converges weakly to the invariant distribution of the original dynamical system. Conditions for when this is in fact true were given as a theorem under the hypothesis of bounded total variation first in \cite{li1976finite}.

Now we pursue other, perhaps more favorable density estimators of the transfer operator, notably \gls{kde}.

\subsection{Theory of \gls{kde}}\label{sec:KDE_theory}

Another major category of data-driven nonparametric density estimators is the \gls{kde}.  It is an estimator based on mixing simple densities.  These are defined in terms of a kernel function, ${\cal K}$, which is itself a real density function.  For single-variate data, ${\cal K}:{\mathbb R} \rightarrow {\mathbb R}^+$, and given: 1) ${\cal K}(x)$ is use enumerate environment symmetric, 2) $\int_{\mathbb R} {\cal K}(x) dx=1$, 3) $\lim_{x\rightarrow \pm\infty}{\cal K}(x)=0.$  These are sufficient to guarantee that the \gls{kde} estimator built out of convex sums of sampling ${\cal K}$ at data points, itself is a density,
\begin{equation}
    \overline{\rho_{N,\delta}} (x)=\frac{1}{\delta N} \sum_{i=1}^N {\cal K}(\frac{x_i-x}{\delta}),
\end{equation}
where $\delta>0$ is the bandwidth that controls the range or extent of influence of a given data point $x_i$ and is a primary parameter choice, just as was the bin size for the histogram method. 

There are several popular kernels, including notably, uniform, triangular, Epanechnikov, Gaussian, and quadratic kernels \cite{silverman_1999,sheather1991reliable, petersen1989ergodic}. These may be chosen with specific properties in mind, such as a kernel with compact support, or otherwise, \cref{tnormal} truncated distributions in general. If $\mathcal{K}$ is given with compact support, then $\mathcal{K}(x)=0$ for values of $x$ lying outside the support. Throughout this article, we employ the Gaussian kernel for numerical demonstrations.

Similarly, for multivariate data, $\overline{\rho_{N,\Sigma}} (x)=\frac{1}{ N} \sum_{i=1}^N {\cal K}_\Sigma(x_i-x) $, using the common compact ``scaled" kernel notation, and ${\cal K}_\Sigma(z)=|\Sigma^{-1/2}|K(\Sigma^{-1/2} z)$ where the most commonly used Gaussian kernel for $D$-dimensional space is given by $\mathcal{K}(z)=(2\pi)^{-D/2} \exp(-z^Tz/2)$ (here $z^T$ represents vector transpose). The matrix $\Sigma$ serves as the variance-covariance matrix in the case of a Gaussian with mean $x_i$.

A crucial difference is whereas histograms are centered on spatial positions, the location of the bins, and data would occupy those positions, a \gls{kde} is centered only where there is data. This is especially useful saving when considering sparsely sampled data from a distribution with relatively small support, especially in higher dimensions where the curse of dimensionality prohibits covering the space with boxes, many of which may be empty of data if the support of the density is zero (no observations or data points in a region or subset).

To analyze \gls{mse}, we must again state the \gls{bias} and variance of the estimator. The \gls{bias} of the estimator is given by,
\begin{eqnarray}
    \gls{bias}_{kde}(\overline{\rho_{N,\delta}}(x))&=&{\mathbb E}(\frac{1}{\delta N} \sum_{i=1}^N {\cal K}(\frac{x_i-x}{\delta})- \rho_X(x)) \nonumber \\
    &=&\frac{1}{N}\int {\cal K}(\frac{y-x}{\delta})\rho(y)dy-\rho_X(x)=
   \int {\cal K}(z)\rho(x+\delta z)dz - \rho_X(x).
\end{eqnarray}
By substituting a Taylor series, $\rho(x+\delta z)=\rho_X(x)+\delta z \rho_X'(z) +\frac{1}{2} \delta^2 z^2 \rho_X''(x) + \mathcal{O}(\delta^2) $, it follows \cite[Section 6]{scott_2015} that,
\begin{equation}\label{biaskde}
    \gls{bias}_{kde}(\overline{\rho_{N,\delta}}(x))=\frac{c}{2} \delta^2 \rho_X''(z) +\mathcal{O}(\delta^2),
\end{equation}
where $c=\int z^2 {\mathcal K}(z) dz$ is the second moment of the kernel and $\mathcal{O}(.)$ is the Big-O notation that shows the convergence rates .

Analysis of variance follows similarly. It can be determined by the following calculation:

\begin{equation}\label{variancekde}
  \gls{var}_{kde}(\overline{\rho_{N,\delta}}(x)) = \gls{var}(\frac{1}{\delta N} \sum_{i=1}^N {\cal K}(\frac{x_i-x}{\delta}))
    \leq \frac{1}{\delta^2 N}{\mathbb E}( {\mathcal K}^2(\frac{x_i-x}{\delta}))  
 \end{equation}
 \begin{equation}
  = \frac{1}{\delta N} \int {\cal K}^2(y) (\rho_X(x)+\delta y \rho_X'(x) + \mathcal{O}(\delta)) dy=
  \frac{1}{\delta N} (\rho_X(x))\int {\cal K}^2 (y) dy + \mathcal{O}(\delta) 
= \frac{1}{\delta N} \rho_X(x) d + \mathcal{O}(\frac{1}{\delta N}),
\end{equation}
with $d=\int {\cal K}^2(y) dy$.

The  \gls{mse} for the \gls{kde} can be obtained by combining \cref{biaskde,variancekde}. It is expressed as:
\begin{equation}\label{Eq:MSE_KDE}
\gls{mse}_{kde}(x)=\frac{c^2}{4}\delta^4 |\rho_X''(x)|^2 + \frac{d}{\delta N} \rho_X(x) + \mathcal{O}(\delta^4)+\mathcal{O}((\delta N)^{-1}).
\end{equation}
Or, 
\begin{equation}
\gls{mse}= \mathcal{O}(\delta^4)+\mathcal{O}((\delta N)^{-1})
\end{equation} moderates the \gls{mse} relative to bandwidth $\delta$ choice. Utilizing \cref{Eq:MSE_KDE} and its derivation process discussed earlier, it can be observed that the bias-variance trade-off in \gls{kde} can be explained by treating the bandwidth $\delta$ as a parameter, where bias dominates for larger values of $\delta>0$ that is proportional to $\rho_X''(x)$ (i.e., curvature), and variance dominates for smaller values of $\delta$ that play the role of bandwidth selection.

\subsection{Optimal \gls{mse}}\label{sec:opt}

The choice of bandwidth tailored to a given data set size is the key question in using a given nonparametric estimator. Despite the fact that the considerations for both histogram and kernel density estimation encompass unknown constants that rely on $\rho_X(x)$ or its derivatives and are beyond our grasp due to the absence of knowledge about $\rho$ (as we may only possess data and not the true distribution), we are only able to make conclusions based on the size of the dataset and the bandwidth utilized. 
\begin{equation}
    \gls{mse}_{hist}(\overline{\rho_{N,K}}(x)) ~\sim \mathcal{O}(\frac{1}{K^2})+ \mathcal{O}(\frac{K}{N}),
\end{equation}
but for \gls{kde},
\begin{equation}
    \gls{mse}_{\gls{kde}}(\overline{\rho_{N,K}}(x)) ~\sim \mathcal{O}(\delta^4) + \mathcal{O}(\frac{1}{\delta N}),
\end{equation}
each balances large \gls{bias} when the bandwidth is too large, versus large variance when the bandwidth times data set size is too small, but at different rates. The asymptotic mean square error can be shown \cite{silverman_1999,scott_2015}  to be optimal when, 
\begin{equation}\label{bandwidth1}
    \delta_{opt;KDE}=\frac{C}{N^{1/5}}, 
\end{equation}
where $C$ is a constant related to the unknown density function, $C=\frac{4 \rho(x) d}{c^2|\rho''(x)|^2} $.
Similarly, for histograms, an optimal bandwidth selection is described by,
\begin{equation}
    K_{opt;hist}=(\frac{N C_1^2}{\rho(\tilde{x})})^{\frac{1}{3}},
\end{equation}

and note that bandwidth/bin-width for a histogram is considered to be as $1/K$.  So we see that asymptotically, cubic versus quintic scaling and the \gls{kde} may be better when optimal bandwidth is used, but in practice, that also depends on the constants, and one depends largely on the $\rho_X$ and the other also on $\rho_X''$.   The most relevant factor to consider when selecting a method is the \gls{mse}. For \gls{kde} with the optimal bandwidth, the \gls{mse} can be expressed as:
\begin{equation}
\gls{mse}_{\delta_{opt;kde}}(\overline{\rho_{\delta, N}}(x))= \mathcal{O}\left(\frac{1}{N^{\frac{4}{5}}}\right).
\end{equation}
On the other hand, for \gls{hde} with the optimal bin size, the \gls{mse} is given by:
\begin{equation}
\gls{mse}_{K_{opt;hist}}(\overline{\rho_{\delta, N}}(x))= \mathcal{O}\left(\frac{1}{N^{\frac{2}{3}}}\right).
\end{equation}
These expressions describe the asymptotic behavior of the \gls{mse} for each method, indicating the rate at which the error decreases as the sample size N increases. It is important to note that \gls{kde} can achieve the expected tolerance level with a smaller sample size when using optimal parameter values compared to \gls{hde}. However, the best we can do is select a parameter value by heuristic rule-of-thumb approach in practice, since we will not know $\rho_X$, to inspect the scaling.  Beyond 1-dimensional density estimation, multivariate \gls{kde} has a slower bandwidth rate \cite{scott_2015} ,
\begin{equation}
        \delta_{opt;KDE}=\frac{C}{N^{\frac{1}{4+D}}}, 
\end{equation}
in $D\geq 1$ dimensions.  For example, the density estimation problem associated with $\rho_X(x)$ is $D=1$ for a transfer operator of the logistic map, but the joint density $\rho_{X'X}(x',x)$ is $D=2$ for the same.

One may utilize either a heuristic rule-of-thumb approach or a systematic optimization method \cite{silverman_1999} to determine the optimal parameter values for density estimation. In this demonstration, as we possess precise knowledge of the density function, we will employ the \gls{ub} to choose the optimal parameter values.

\section{Results and Discussion}\label{sec:rest}
In this section, we focus on estimating the Frobenius-Perron operator using the previously discussed density estimation methods. In other words, we calculate the $\gls{eFP}$ matrix in \cref{est4} by density estimation methods. We use the logistic map example to demonstrate the results. In this demonstration, uniformly distributed $N=10^6$ initial conditions were used and evolved using a logistic map $x'=4x(1-x)$ for a relatively long time which was used to approximate the invariant density $\rho(x)=\frac{1}{\pi\sqrt{x(1-x)}}$. Now our goal is to estimate the Frobenius-Perron operator by evaluating the probability density function $\rho(x'|x)=\frac{\rho(x,x')}{\rho(x)}$. For this calculation, we estimated the $\rho(x)$ and the joint probability density $\rho(x,x')$ using the data through density estimation methods. We analyze the estimation of the $\rho(x)$ by the density estimation methods in detail and compare it to the theoretical explanation in the section \ref{Sec:TheoryDen}. Then we demonstrate the estimation $\gls{eFP}$ matrix of the Frobenius-Perron operator by discretized probability density function $\rho(x'|x)$.

 In this section, numerical calculations of the \gls{kde} use the ``ksdensity" MATLAB function, while \gls{hde} utilizes the ``histogram" MATLAB function. . The ``ksdensity" function employs the following expression for the bounded domain $[L,U]$ with kernel $\mathcal{K}$:
\begin{equation*}
\overline{\rho_{N,\delta}}(x) = \frac{1}{N\delta} \sum_{i=1}^{N} \left[ \mathcal{K}\left(\frac{x-x_{i}^{-}}{\delta}\right) + \mathcal{K}\left(\frac{x-x_i}{\delta}\right) +  \mathcal{K} \left(\frac{x-x_{i}^+}{\delta}\right) \right] \quad \text{for } L \leq x \leq U,
\end{equation*}
where $x_{i}^- = 2L - x_i$, $x_{i}^+ = 2U - x_i$, and $x_i$ represents the $i^{\text{th}}$ sample data  \cite{MATLAB:2022}.

\subsection{\gls{hde} vs \gls{kde} for Estimating Invariant Density of Logistic Map}

\begin{figure}[htbp]
\centering
\includegraphics[width=0.6
\textwidth]{ 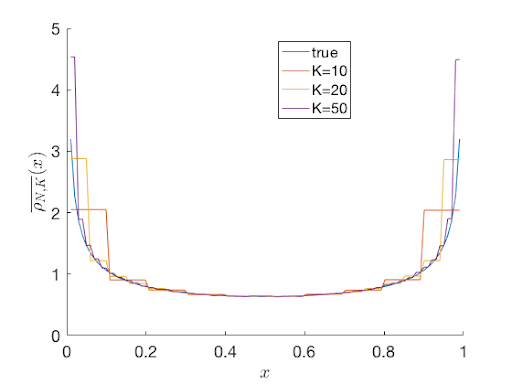}
\caption{ The figure shows the results of histogram-based density estimation, $\overline{\rho_{N,K}}$, for the invariant density, $\rho$, of the logistic map. The impact of the number of bins, $K$, on the estimation $\overline{\rho_{N,K}}$ is illustrated. The sample size used for estimation is $N=10^6$. The figure provides insights into how the choice of the number of bins affects the accuracy and bias of the density estimation. Specifically, it reveals how the choice of $K$ impacts the smoothness and variability of the estimated density. }
\label{fig:histEstK}
\end{figure}

\begin{figure}[htbp]
\centering
\includegraphics[width=0.45
\textwidth]{ 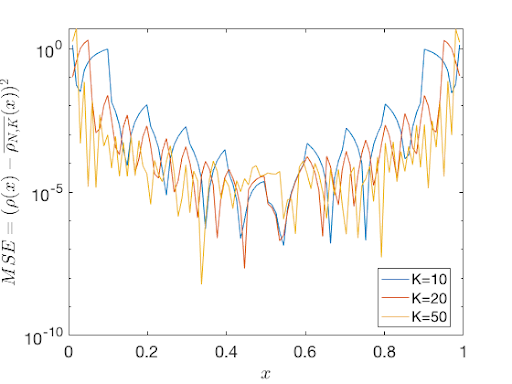}
\includegraphics[width=0.45
\textwidth]{ 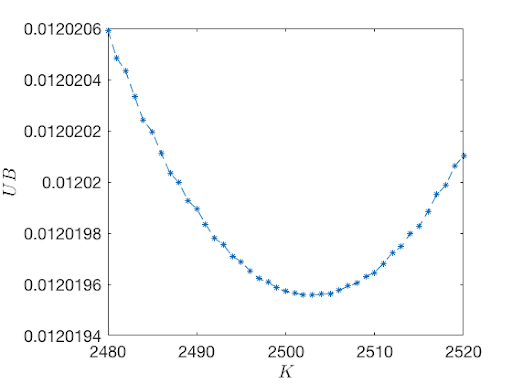}
\caption{(Left) The \gls{mse} of the estimated density function on a uniformly spaced grid of $100$ points in the interval $[0.01,0.99]$, as a function of the bin size $K$. The right figure shows the behavior of the \gls{ub} with the bin size for the logistic map example. The optimal \gls{mse} can be achieved at a bin size of $K_{opt}=2503$. }
\label{fig:histMSE_UB}
\end{figure}

Histogram estimation is based on the number of samples ($N$) and the number of bins ($K$) (See details in section \ref{Sec:TheoryDen}. Here, we demonstrate the effect of the bin size. The estimation of the invariant density $\rho(x)$ by the histogram method is denoted by $\overline{\rho_{N,K}}$. \cref{fig:histEstK} shows the changes in the estimation with parameter $K$. Since the true density is known, the \gls{mse} and the \gls{ub} of the \gls{mse} can be calculated from \cref{msehist}. Note that the notations 
\begin{align*}
    \gls{mse} &=(\rho(x)-\overline{\rho_{N,K}}(x))^2 \\
    \gls{ub} & = C_1^2 \frac{1}{K^2}+  \frac{\rho(\hat{x}) }{N} K+ \frac{\rho^2(\hat{x})}{N}
\end{align*}
are from \cref{msehist} and depend on the true density function $\rho(x)$. Additionally, through the examination of the \gls{ub} and \gls{mse}, we can determine the optimal bin size, $K_{opt;hist}$, as illustrated in \cref{fig:histMSE_UB} and discussed in \cref{sec:opt}.


\begin{figure}[htbp]
\centering
\includegraphics[width=0.6
\textwidth]{ 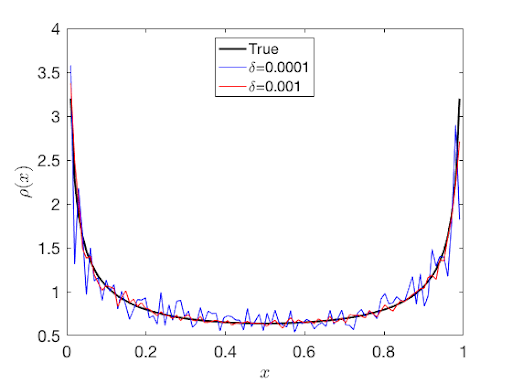}
\caption{ The figure illustrates the \gls{kde} of the invariant density $\rho$ of the logistic map. It depicts how the bandwidth parameter $\delta$ affects the \gls{kde} with sample size $N=10^6$. Specifically, it reveals how the choice of $\delta$ impacts the smoothness and variability of the estimated density.}
\label{Fig:KDE_logi}
\end{figure}

\begin{figure}[htbp]
\centering
\includegraphics[width=0.45
\textwidth]{ 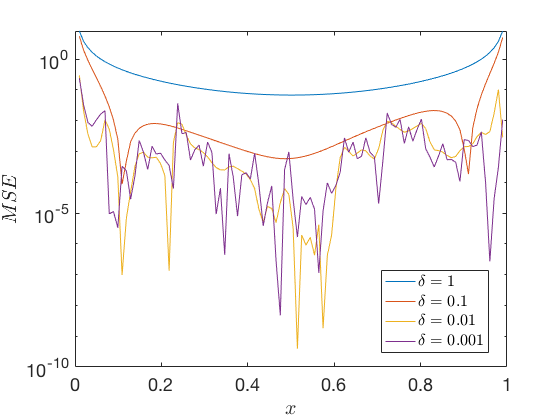}
\includegraphics[width=0.45
\textwidth]{ 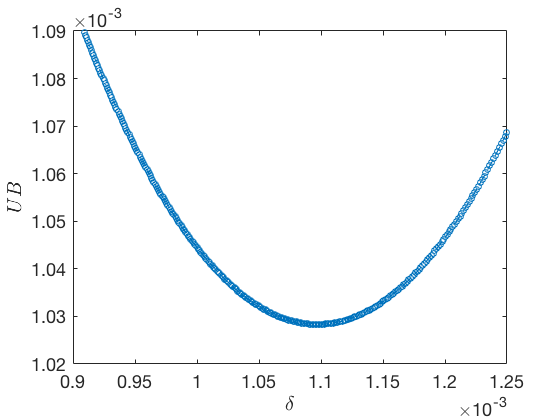}
\caption{(Left) Displays of the \gls{mse} of the estimated density function on a uniformly spaced grid of $100$ points over the interval $[0.01, 0.99]$. It also shows how the \gls{mse} changes with the bandwidth parameter $\delta$ for the \gls{kde}. The right figure presents the analysis of the \gls{ub} and how it varies with the bandwidth parameter for the logistic map example. The optimal bandwidth can be identified at around $\delta_{opt}=0.0011$. }
\label{fig:KDEMSE_UB}
\end{figure}

As we discussed in \cref{sec:KDE_theory}, \gls{kde} is based on the number of data points ($N$) and the kernel bandwidth ($\delta$). We now numerically demonstrate (employing the Gaussian kernel for these demonstrations) the effect of the bandwidth. Notice that, a change in the bandwidth will result in a change in the \gls{mse} (see \cref{Fig:KDE_logi}).  The \gls{ub} of \gls{mse} for the \gls{kde} is given in \cref{Eq:MSE_KDE} and the following results are calculated by \cref{Eq:MSE_KDE}. Error analysis and the optimal \gls{mse} is demonstrated in \cref{fig:KDEMSE_UB}. Furthermore, note that the optimal \gls{mse} can be achieved with a bandwidth of approximately $\delta_{opt;KDE}=0.0011$.

\begin{figure}[htbp]
\centering
\includegraphics[width=0.45
\textwidth]{ 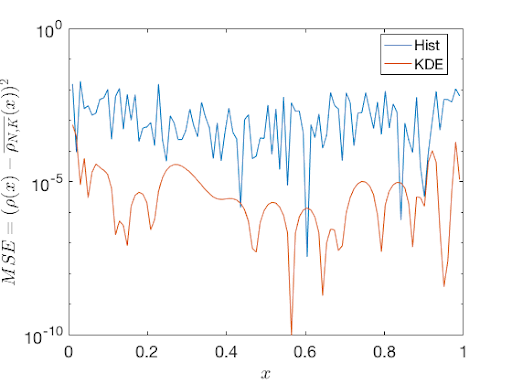}
\caption{This figure presents a comparison of the \gls{mse} of the histogram density estimation  and \gls{kde}, both with their respective optimal parameter values $K_{opt}=2503$ and $\delta_{opt}=0.0011$. The sample data used is the invariant density of the logistic map, with a sample size of $N=10^6$ and equally spaced grid points spanning the interval $[0.01,0.99]$.}
\label{Fig:HistVsKDE}
\end{figure}

Due to the unboundedness of the density function, both estimation methods have higher estimation errors closer to the endpoints of the interval $[0,1]$. In general, density estimation  has issues when estimating an unbounded probability density with finite support. Despite this limitation, the \gls{kde} method tends to have a lower overall error when compared to the histogram method using finite data. \cref{Fig:HistVsKDE} demonstrates that \gls{mse} for \gls{kde} is comparatively lower than the \gls{mse} for histogram method with their optimal parameter values.

\subsection{Estimating the Frobenius-Perron operator} 
In the first part of this section, we further demonstrate the application of our approach using the logistic map example. Here, a numerical investigation of the theoretical discussion presented in \cref{Sec:BayInter} is presented. Through the utilization of the popular classical Ulam-Galerkin method as a histogram estimation of the conditional probability density function $\rho(x'|x)$, we establish a reference for comparison of  the estimation of the Frobenius-Perron operator. Our work highlights the fact that any density estimation method can be employed to accomplish this extension. 

To validate the efficacy of our proposed method in estimating the Frobenius-Perron operator, we examine an additional example. This examination serves to reinforce the robustness and reliability of our approach. By presenting the \gls{kde} as an alternative density estimation method to the histogram technique, we contribute to the existing body of knowledge and offer a comprehensive analysis of the subject matter.

For both examples, we utilize square bins with dimensions $K \times K$, where $K=200$ for the \gls{hde} and a bandwidth of $\delta=0.01$ for the \gls{kde} with a Gaussian kernel.

\subsubsection{Example: Logistic Map}

In this subsection, we present a comprehensive example using the logistic map to demonstrate the practical application of our approach. In the analysis, we consider a dataset $(X,X')$ consisting of $N=10^6$ samples. Here, $X$ represents the transformed data obtained by applying the logistic map for two iterations, starting from data uniformly distributed in the range $[0,1]$. Similarly, $X'$ represents the transformed data obtained after three iterations of the logistic map. 
\begin{figure}[ht!]
\centering
\includegraphics[width=0.45
\textwidth]{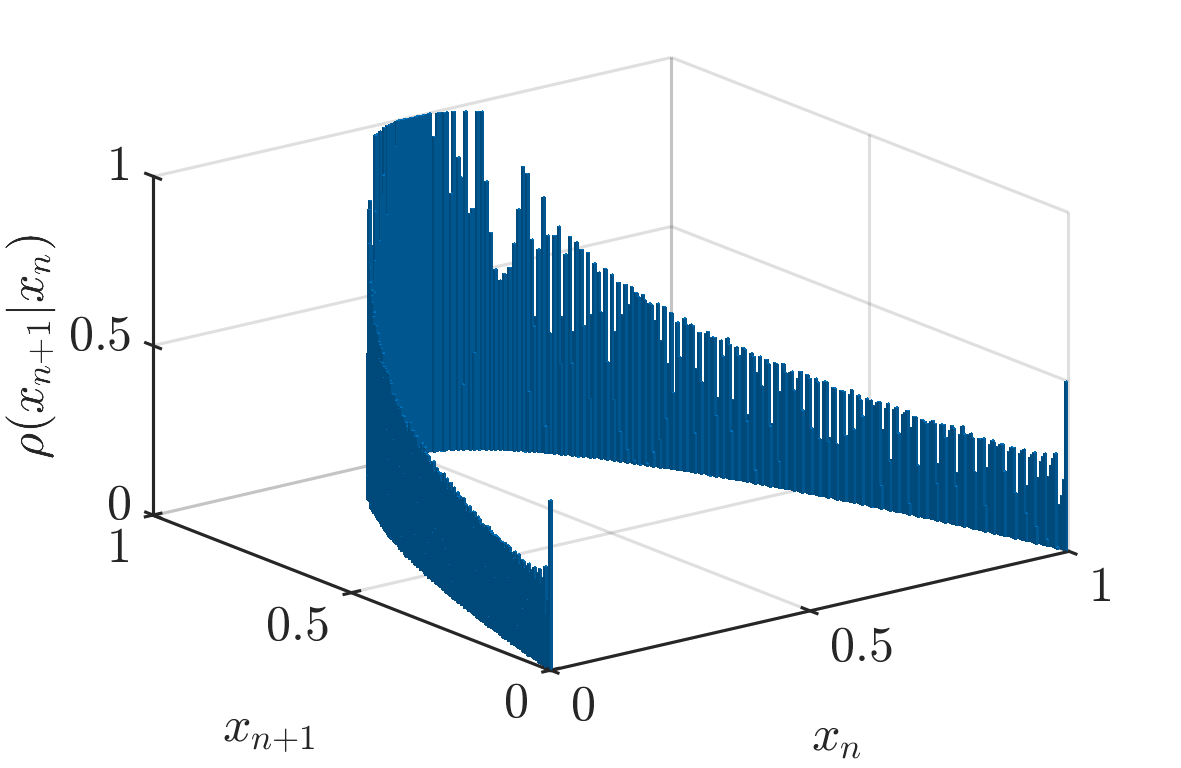}
\includegraphics[width=0.45
\textwidth]{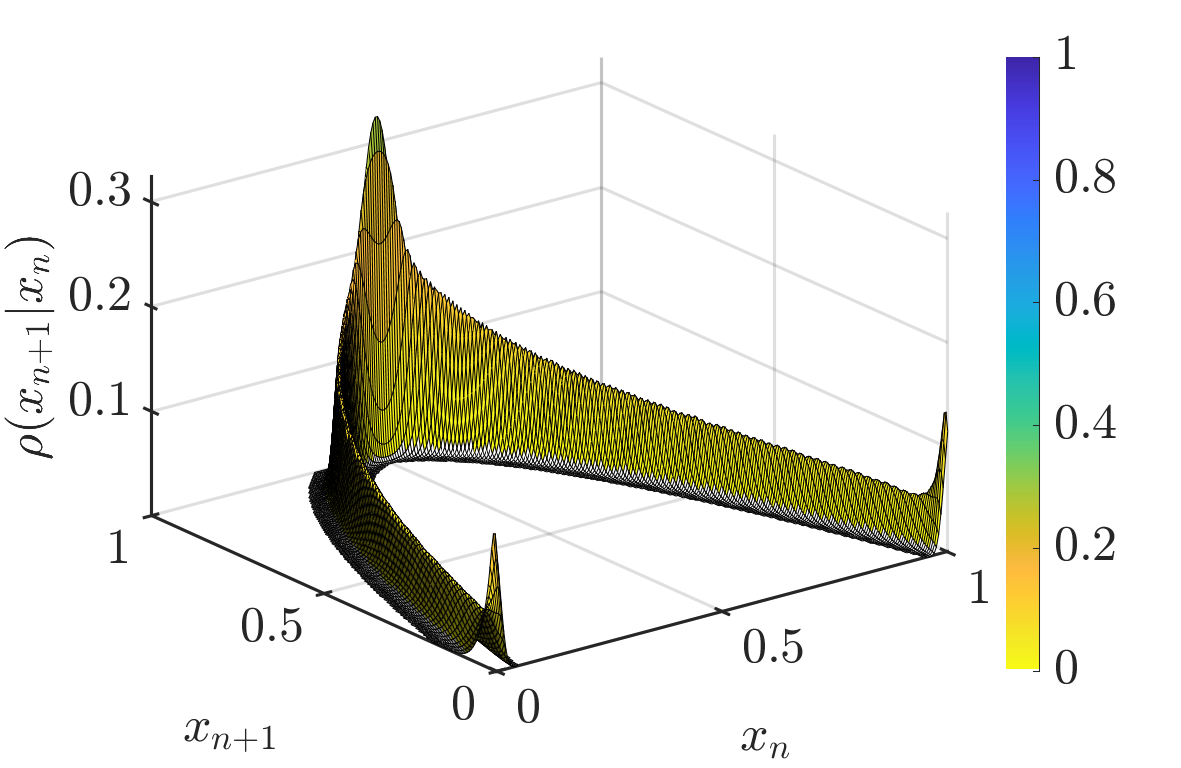}
\caption{ This figure illustrates the estimated probability density function, denoted as $\rho(x_{n+1}|x_n)$, using two distinct density estimation techniques: the \gls{hde} with square bins $K\times K$, where $K=200$ (left), and the \gls{kde} employing a Gaussian Kernel with a bandwidth of $\delta = 0.01$ (right).}
\label{fig:PxpGxDens}
\end{figure}
\begin{figure}[ht!]
\centering
\includegraphics[width=0.45
\textwidth]{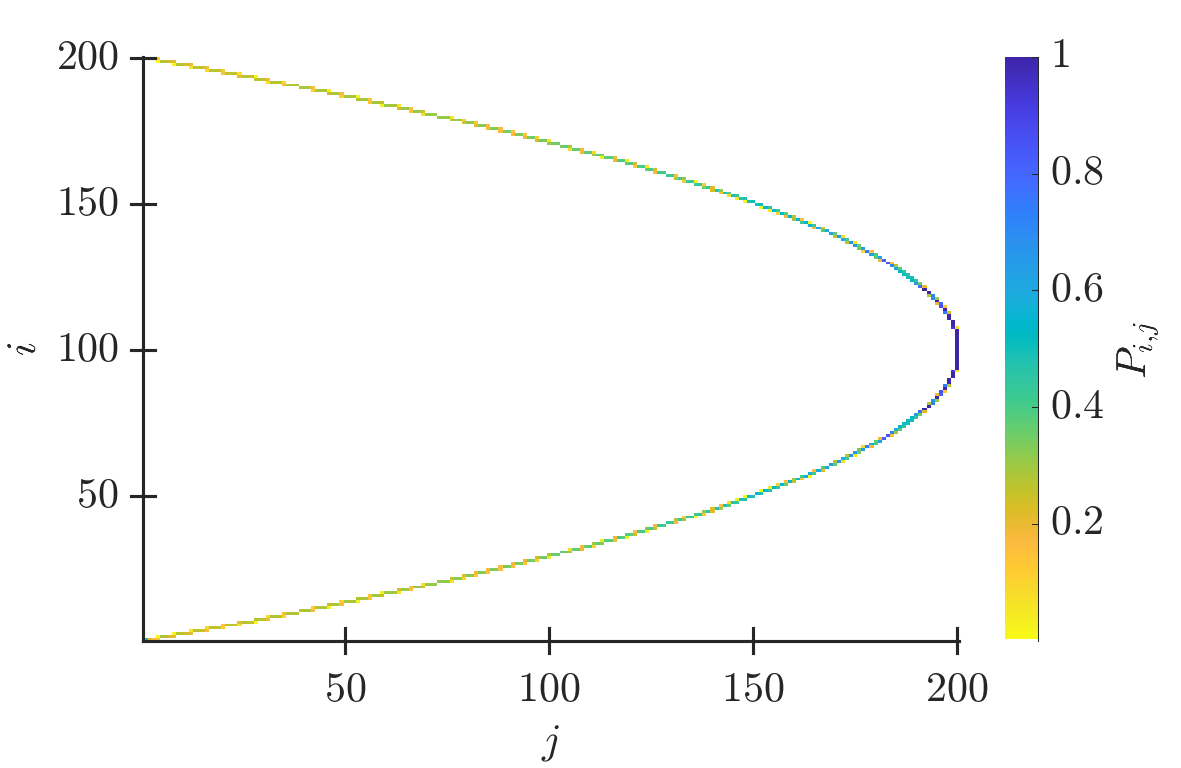}
\includegraphics[width=0.45
\textwidth]{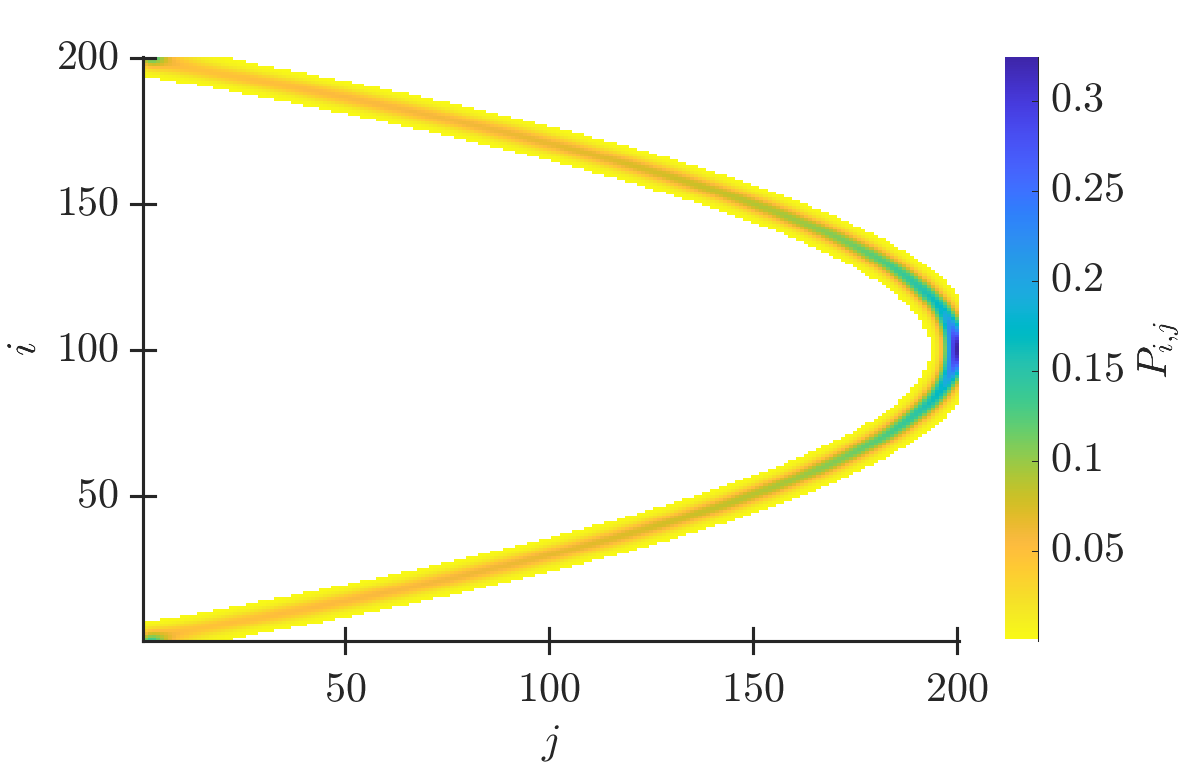}
\caption{ This figure presents estimates of elements of the matrix $\gls{eFP}$, which approximates the Frobenius-Perron operator in a finite domain, obtained using two density estimation techniques: the \gls{hde}(left) and the \gls{kde} (with Gaussian Kernel) (right).}
\label{fig:Pmatrix}
\end{figure}

\cref{fig:PxpGxDens} showcases the estimated probability density function $\rho(x_{n+1}|x_n)$ using two density estimation techniques: the \gls{hde} (left) and the \gls{kde}(right). To approximate the Frobenius-Perron operator within a finite domain, we derive estimates of the matrix $\gls{eFP}$. The left plot in \cref{fig:Pmatrix} illustrates the $\gls{eFP}$ matrix obtained using the \gls{hde}, while the right plot demonstrates the results obtained through \gls{kde}. In this case, we normalize the rows of the matrix to ensure it is stochastic. 

\begin{figure}[ht!]
\centering
\includegraphics[width=0.45
\textwidth]{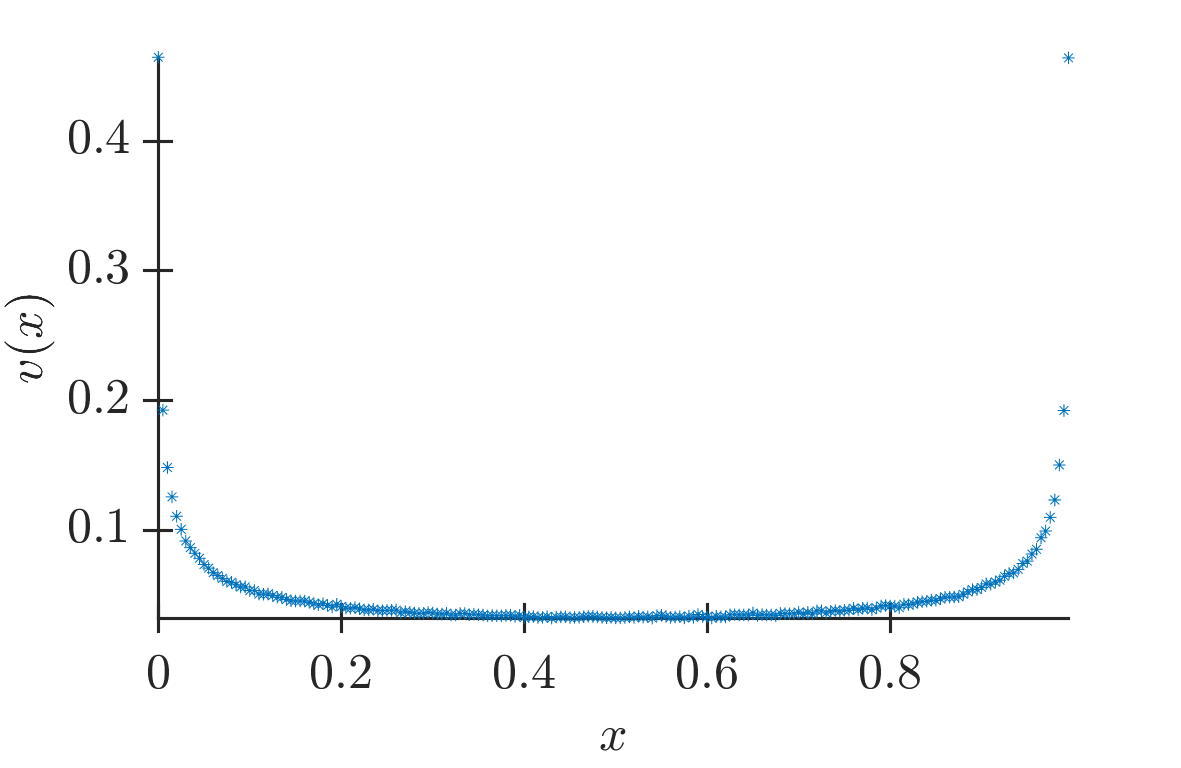}
\includegraphics[width=0.45
\textwidth]{ 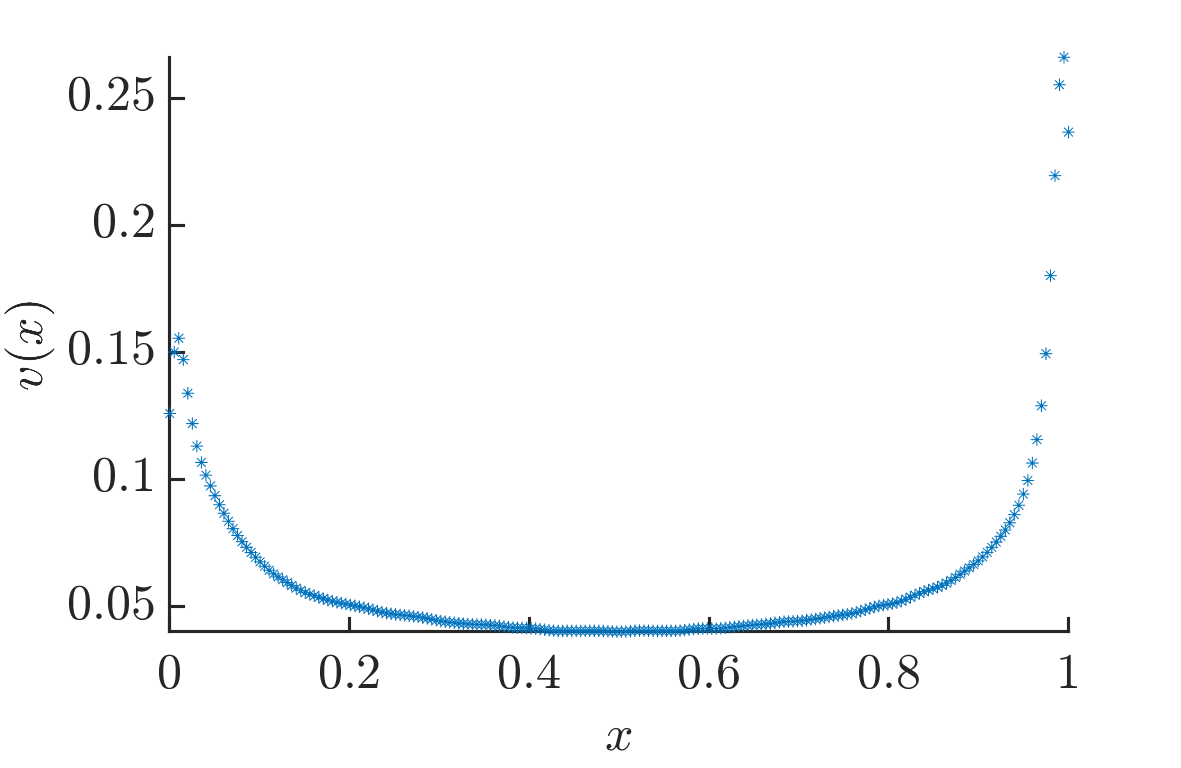}
\caption{The left eigenvector of matrix $\gls{eFP}$ is displayed, which corresponds to the eigenvalue $1$ of the matrix $\gls{eFP}$ shown in \cref{fig:Pmatrix}. The left eigenvectors of $\gls{eFP}$ estimated using two density estimation techniques are shown in the figure: the histogram method (left) and the kernel density estimation (KDE) (right).}
\label{fig:lestEigs}
\end{figure}

To estimate the invariant density of the logistic map, we examine the left eigenvector of the matrix $\gls{eFP}$ corresponding to the eigenvalue $1$. \cref{fig:lestEigs} depicts the left eigenvectors estimated using both the \gls{hde}(left) and the \gls{kde}(right). These findings provide compelling evidence supporting the validity and effectiveness of our approach in estimating the Frobenius-Perron operator by interpreting the classical Ulam-Galerkin approach as a density estimation technique.

\subsubsection{Example: Markov Map}
In this subsection, we present additional evidence to substantiate the validity of our proposed approach, which extends the conventional Ulam-Galerkin method by incorporating density estimation techniques. To demonstrate the efficacy of our approach, we employ a Markov map as described by \citeauthor{bollt2013applied} \cite[Section 4.3]{bollt2013applied}. This specific selection of the Markov map holds notable advantages, primarily due to the analytical derivability of the eigenvector associated with the Frobenius-Perron operator for this particular map, as expounded in the aforementioned book \cite[pages 76-79]{bollt2013applied}. The Markov map $g: [0,4] \rightarrow [0,4]$ can be written as follows:

\begin{equation}
    g= \begin{cases}
  3x & \text{if }\ x < 1, \\
  -x + 4 & \text{if }\ 1 \leq x < 2, \\
  2x - 2 & \text{if }\ 2 \leq x < 3, \\
  -4x + 16 & \text{else }\ x \geq 3
\end{cases}.
\end{equation}

The transition matrix can be represented by:
\begin{equation}
\setlength{\arraycolsep}{15pt}
    A= \begin{pmatrix}
1/3 & 1/3 & 1/3 & 0 \\
0 & 0 & 1 & 0 \\
0 & 0 & 1/2 & 1/2 \\
1/4 & 1/4 & 1/4 & 1/4
\end{pmatrix}.
\end{equation}
In this scenario, the Frobenius-Perron operator can be represented as an operator with finite rank when applying Galerkin's method using a finite number of terms. Moreover, the approximation becomes perfectly accurate when employing a Markov partition and the corresponding basis functions \cite[pages 78]{bollt2013applied}. Hence, the steady state of the operator is given by the left eigenvector:

\begin{equation}
    v= \frac{1}{\sqrt{226}}\begin{pmatrix}
3 \\
3\\
12 \\
8
\end{pmatrix}
\end{equation}
of the transition matrix $A$ corresponding to the largest eigenvalue, which is $1$. 

Now we will demonstrate our density estimation approach by estimating this steady state using the dataset $(X,X')$ of size $N=10^6$. Here, $X$ represents the transformed data obtained by applying the Markov map for two iterations, starting from data uniformly distributed in the range $[0,4]$. Similarly, $X'$ represents the transformed data after three iterations. 

\begin{figure}[htbp]
\centering
\includegraphics[width=0.45
\textwidth]{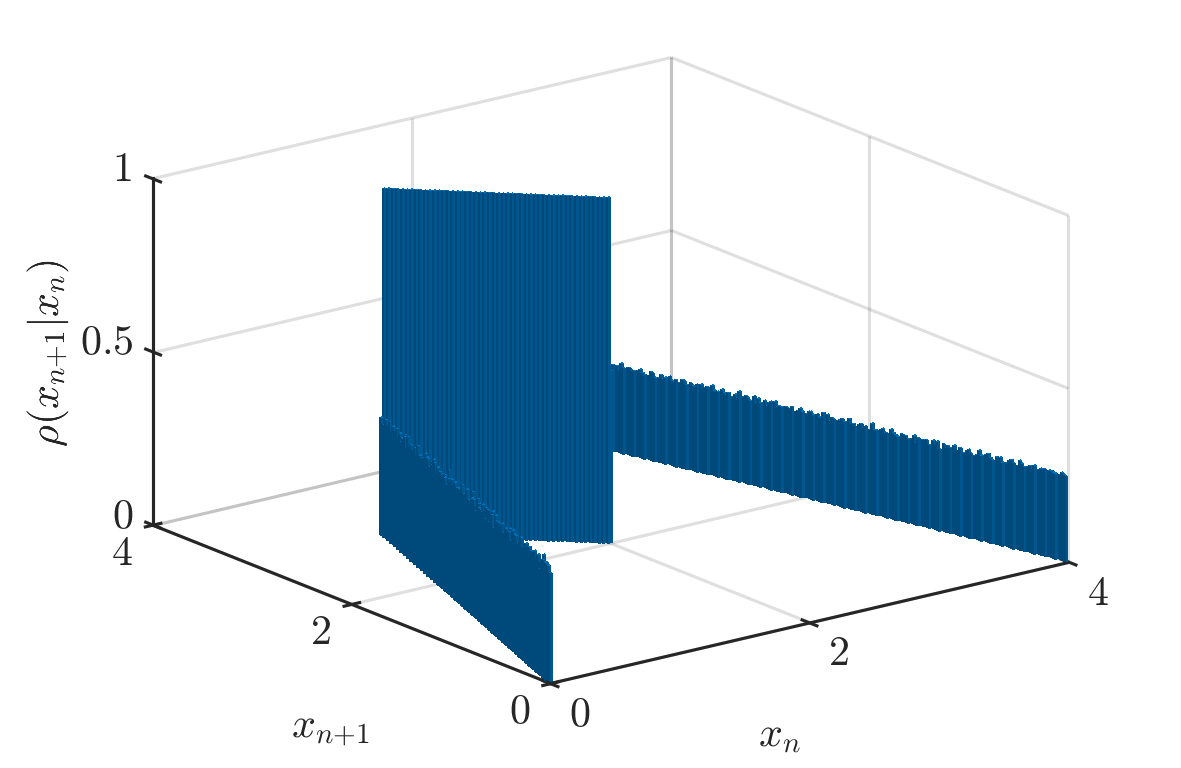}
\includegraphics[width=0.45
\textwidth]{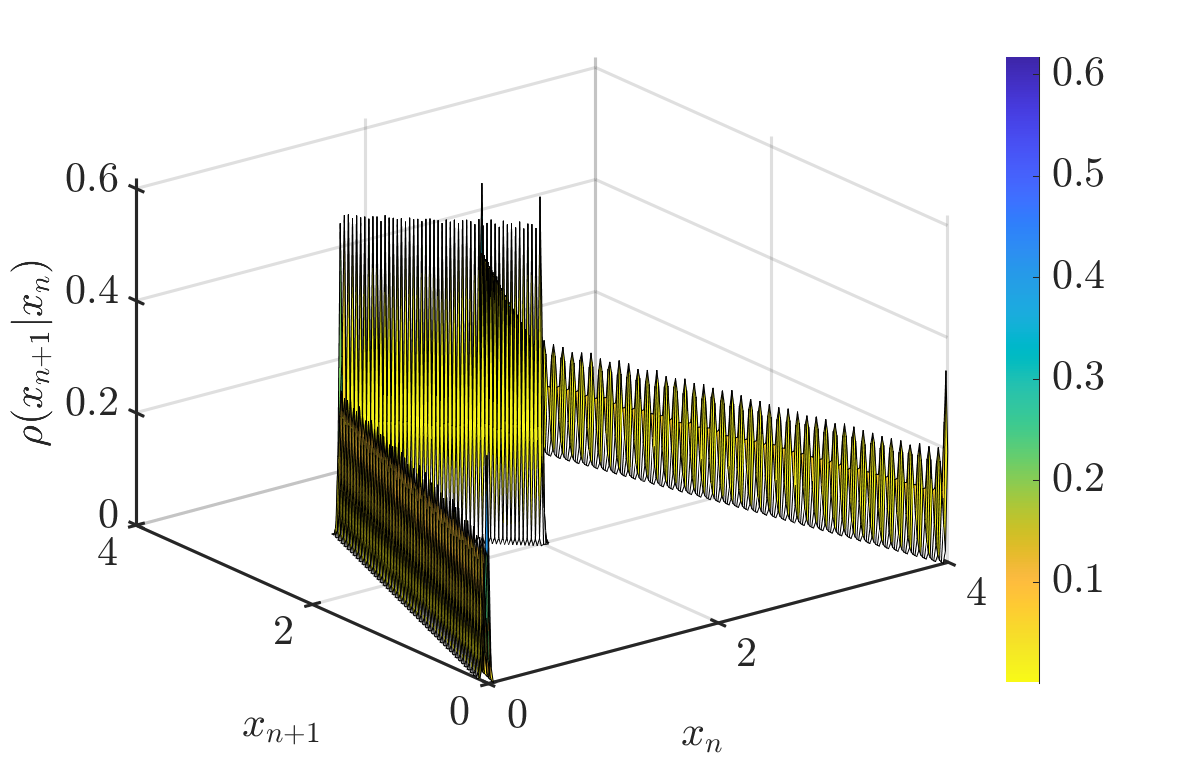}
\caption{ This figure illustrates the estimated probability density function, denoted as $\rho(x_{n+1}|x_n)$, using two distinct density estimation techniques: the \gls{hde} with a bin size of $K=200$ (left), and the \gls{kde} employing a Gaussian Kernel with a bandwidth of $\delta = 0.01$ (right).}
\label{fig:PxpGxDensMmap}
\end{figure}

\begin{figure}[htbp]
\centering
\includegraphics[width=0.45
\textwidth]{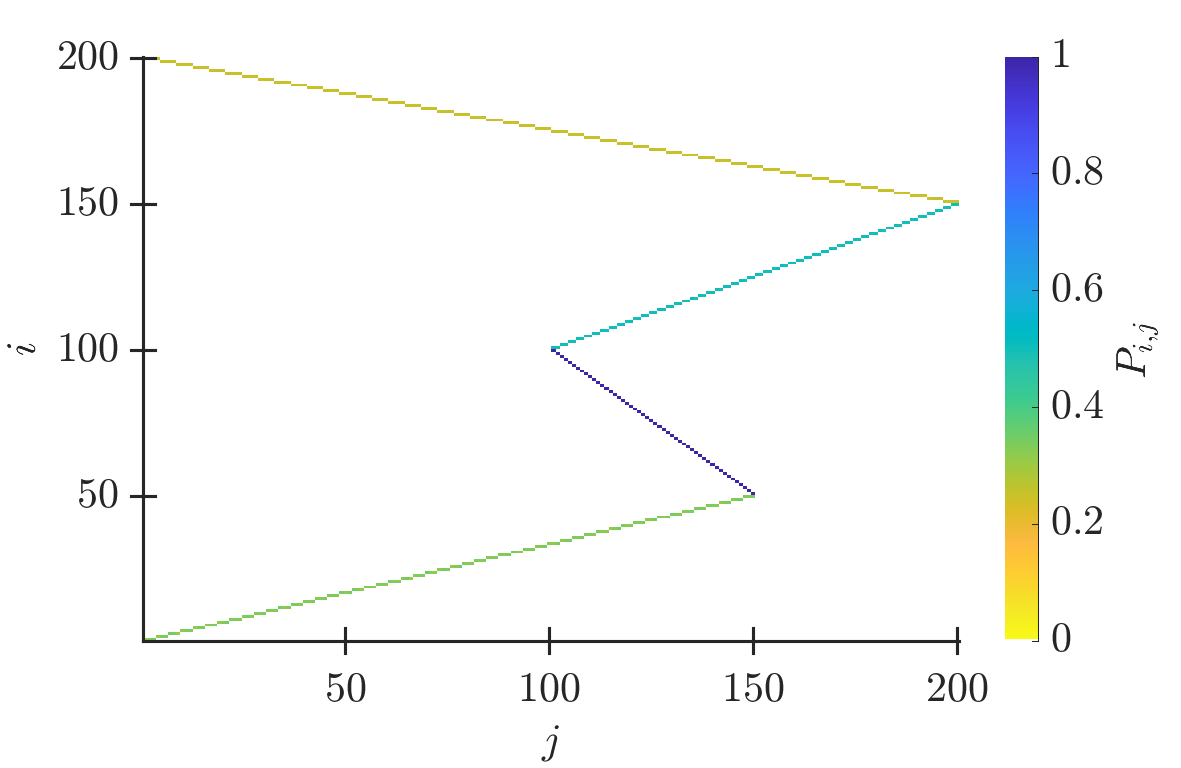}
\includegraphics[width=0.45
\textwidth]{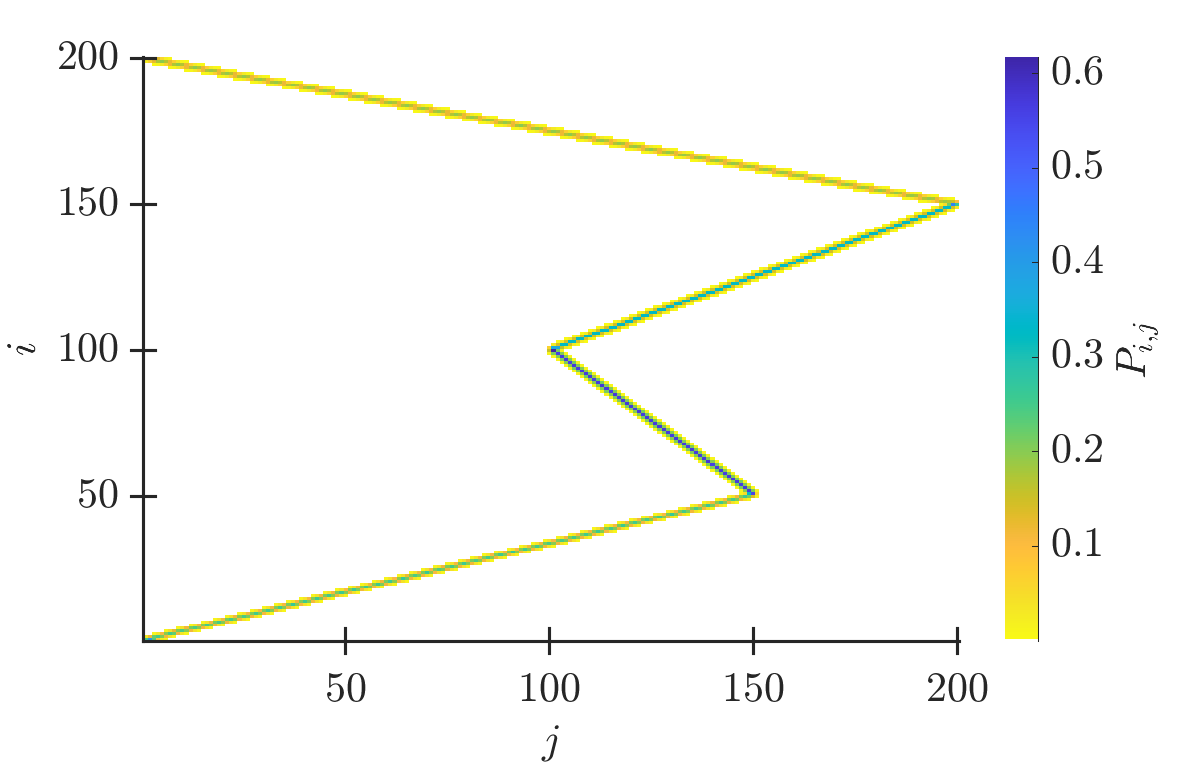}
\caption{ This figure presents estimates of the matrix $\gls{eFP}$, which approximates the Frobenius-Perron operator in a finite domain, obtained using two density estimation techniques: the histogram method (left) and the \gls{kde} (with Gaussian Kernel) (right).  The matrix $\gls{eFP}$ has dimensions $K\times K$, where $K=200$ is the number of bins or grid points used in the density estimation. }
\label{fig:PmatrixMmap}
\end{figure}

Similar to the example of the logistic map, we use the number of partitions $K=200$ for our analysis. In the case of the \gls{hde} method, we employ square bins with dimensions $K\times K$. For the \gls{kde} method, we utilize a bandwidth of $\delta=0.01$ along with a Gaussian kernel.  \cref{fig:PxpGxDensMmap} illustrates the estimation of the conditional probability density $\rho(x_{n+1}|x_n)$ using the \gls{hde} method (left) and the \gls{kde} method (right). The estimated conditional densities are then used to derive the matrix $\gls{eFP}_{K\times K}$, which is shown in \cref{fig:PmatrixMmap}. In this case, we normalize the rows of the matrix to ensure it is stochastic.

\begin{figure}[htbp]
\centering
\includegraphics[width=0.45
\textwidth]{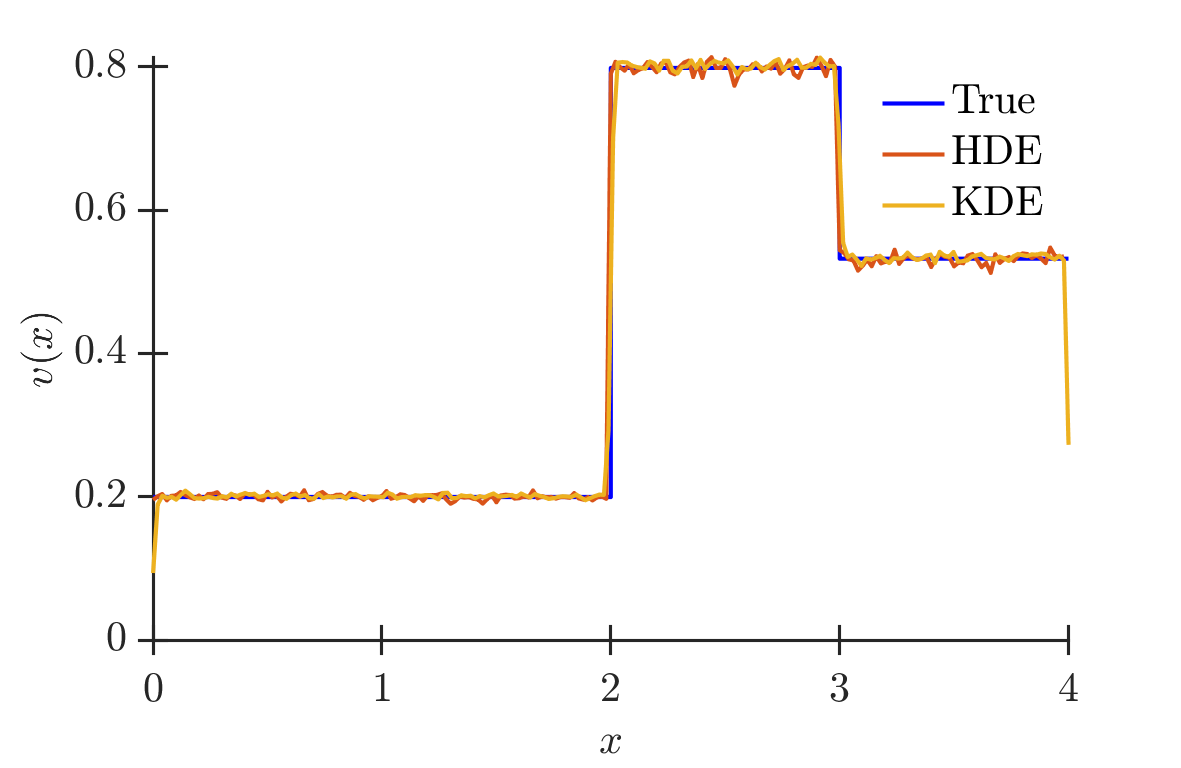}
\includegraphics[width=0.45
\textwidth]{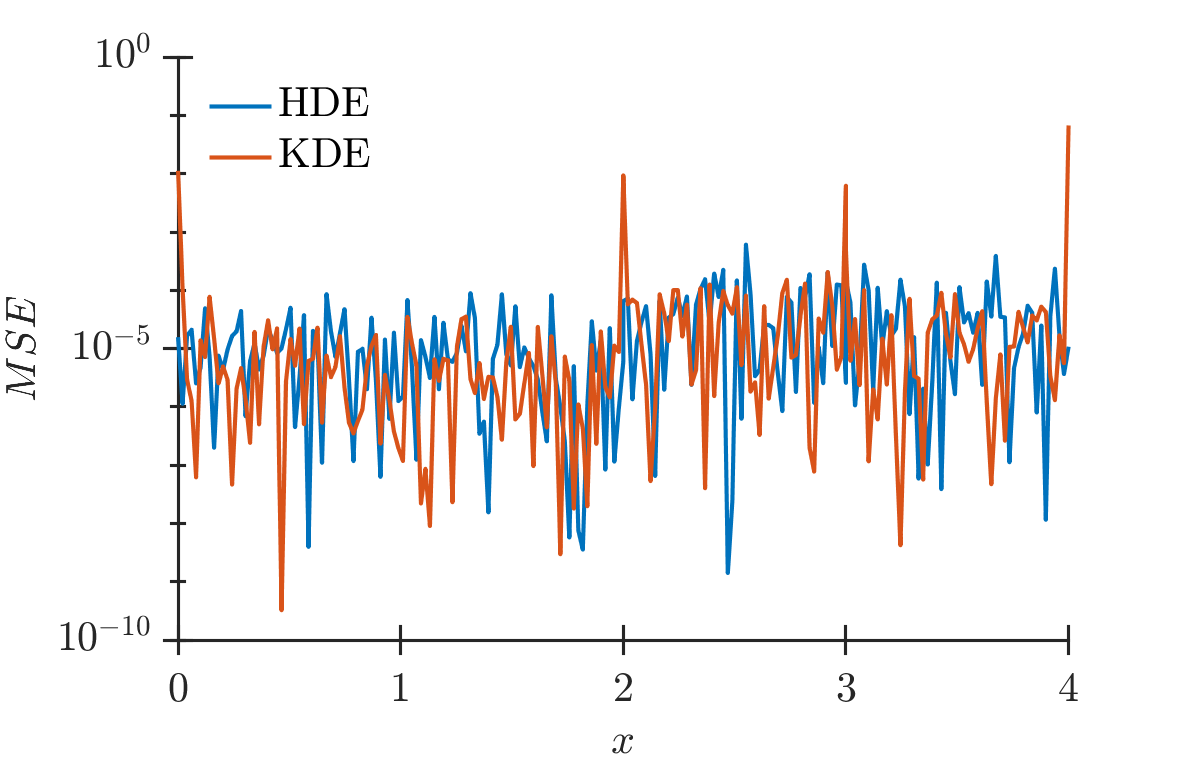}
\caption{The figure analyzes the left eigenvector of the matrix $\gls{eFP}$, which corresponds to the eigenvalue $1$ of the matrix $\gls{eFP}$ shown in Figure \ref{fig:PmatrixMmap}. The left figure compares the true eigenvector with the estimated eigenvectors of $\gls{eFP}$. The right figure illustrates the \gls{mse} of the estimated eigenvectors. }
\label{fig:lestEigsMmap}
\end{figure}

Finally, we compare the invariant density with the estimated eigenvectors of the $\gls{eFP}$ matrix, which is estimated using density estimation methods (\cref{fig:lestEigsMmap}). To compare the estimated normalized eigenvector $\hat{v}(x)$ with the true value $v(x)$, we multiply $\hat{v}(x)$ by the constant $\gamma = \frac{\sqrt{3^2 \times 100 + 12^2 \times 50 + 8^2 \times 50}}{\sqrt{226}}$. The left plot in \cref{fig:lestEigsMmap} compares the eigenvectors derived from the \gls{hde} and \gls{kde} methods with the true eigenvector. The right plot in the figure displays the \gls{mse} of the eigenvectors for the \gls{hde} and \gls{kde} methods. These comparisons demonstrate the validity of interpreting the Ulam-Galerkin method as a density estimation technique. It is worth noting that while the overall error for estimating the invariant density through the \gls{kde} method is relatively low, there are significantly higher errors observed around boundary points and jumps.


\section{Conclusion}
The main contribution of this paper is the probability density viewpoint to the estimating Frobenius–Perron operator that enables us to incorporate the existing rich analysis of statistical density estimation formalism to find an efficient estimator. Furthermore, the theory suggests that the \gls{kde} is more efficient than the histogram methods used in the standard Ulam method. Additionally, this paper discusses a \gls{kde} method to estimate the transition probability that estimates the Frobenius–Perron operator, from empirical time series ensemble data of a dynamical system. To date, the literature mostly used  the Ulam method for estimating the transfer operator but this study offers a more accurate estimation in terms of \gls{mse}, based on \gls{kde}. Our Bayesian interpretation of the Frobenius–Perron operator is important to identify the operator in terms of conditional probability density because it allows us to bring density estimation theory into play. It is shown at the beginning of this article how the Ulam-Galerkin method can be interpreted as a histogram density estimation method. Theory and numerical results have been presented which suggest that \gls{kde} is a better approximation for estimating probability densities. Hence, it is also shown that \gls{kde} may be used for finite approximation for the Frobenius–Perron operator. 

Furthermore, by utilizing data from both the logistic map and the Markov map, we have successfully estimated the invariant density of the map by estimating the eigenvector of the operator using density estimation methods. This provides further validation for our interpretation of the Ulam-Galerkin method as a density estimation technique. Moreover, it opens up opportunities to explore additional density estimation methods in this context. In this article, a comparison of the \gls{hde} and \gls{kde} methods was offered.

Overall, the \gls{kde} method outperformed the \gls{hde} method, as demonstrated in the Markov map example where the \gls{kde} estimates the eigenvalues of the operator more accurately for a majority of the data points. However, it is important to note that the \gls{kde} method performs poorly around boundary points and jumps. Future research can focus on investigating alternative estimation methods to address these issues and improve the accuracy of the density estimation technique in such scenarios.

As a result of conducting this research, we propose the exploration of incorporating various density estimation methods into this field. In particular, it would be fruitful to pursue further research regarding the application of \gls{kde}-based estimation of the Frobenius-Perron operator for high-dimensional maps. Investigating this method in higher dimensions can provide valuable insights and contribute to a more comprehensive analysis of density estimation techniques in this context.

\newpage


 \nonumsection{Acknowledgments} \noindent Erik Bollt gratefully acknowledges funding from the Army Research Office (ARO), the Defense Advanced Research Projects Agency (DARPA), the National Science Foundation (NSF) and National Institutes of Health (NIH) CRNS program, and the Office of Naval Research (ONR) during the period of this work. 

\nonumsection{Code availability} \noindent
The code used in this study is available at \url{https://github.com/sudamphy/FPoperatorDE.git}.

\printunsrtglossary[type=abbreviations]

\bibliographystyle{ws-ijbc}
\typeout{}
\bibliography{sample}

\end{document}